\documentclass[runningheads]{llncs}
\usepackage[T1]{fontenc}
\usepackage{graphicx}
\usepackage{multirow}
\usepackage{diagbox}
\begin{document}
\title{Anomaly Detection by Effectively Leveraging Synthetic Images}
%
%
\author{Sungho Kang\inst{1}\orcidID{0009-0008-5829-4503} \and
Hyunkyu Park\inst{1}\orcidID{0009-0003-4889-1498} \and
Yeonho Lee\inst{1}\orcidID{0009-0006-7331-3501} \and
Hanbyul Lee\inst{2}\orcidID{0009-0007-2713-477X} \and
Mijoo Jeong\inst{3}\orcidID{0009-0006-1081-2311} \and
YeongHyeon Park\inst{4}\orcidID{0000-0002-5686-5543} \and
Injae Lee\inst{1}\orcidID{0009-0007-4730-8036} \and
Juneho Yi\inst{1}\orcidID{0000-0002-9181-4784}}
\authorrunning{S. Kang et al.}
%

\institute{Department of Electrical and Computer Engineering, Sungkyunkwan University, Suwon 16419, Republic of Korea\\
\email{\{sungho369,mjss016,tarazed,injea1291,jhyi\}@skku.edu} \and
Department of Artificial Intelligence, Sungkyunkwan University, Suwon 16419, Republic of Korea\\
\email{byul1748@g.skku.edu} \and
Department of Architecture, Sungkyunkwan University, Suwon 16419, Republic of Korea\\
\email{gorgeousmj2@g.skku.edu}\and
Department of Radiation Physics, The University of Texas MD Anderson Cancer Center, Texas 77030, USA \\
\email{ypark6@mdanderson.org}}
\maketitle              
\vspace{-0.5cm}

\begin{abstract}
Anomaly detection plays a vital role in industrial manufacturing. Due to the scarcity of real defect images, unsupervised approaches that rely solely on normal images have been extensively studied. Recently, diffusion-based generative models brought attention to training data synthesis as an alternative solution. In this work, we focus on a strategy to effectively leverage synthetic images to maximize the anomaly detection performance.
Previous synthesis strategies are broadly categorized into two groups, presenting a clear trade-off. Rule-based synthesis, such as injecting noise or pasting patches, is cost-effective but often fails to produce realistic defect images. On the other hand, generative model-based synthesis can create high-quality defect images but requires substantial cost. To address this problem, we propose a novel framework that leverages a pre-trained text-guided image-to-image translation model and image retrieval model to efficiently generate synthetic defect images. 
Specifically, the image retrieval model assesses the similarity of the generated images to real normal images and filters out irrelevant outputs, thereby enhancing the quality and relevance of the generated defect images. 
To effectively leverage synthetic images, we also introduce a two-stage training strategy. In this strategy, the model is first pre-trained on a large volume of images from rule-based synthesis and then fine-tuned on a smaller set of high-quality images. This method significantly reduces the cost for data collection while improving the anomaly detection performance. Experiments on the MVTec AD dataset demonstrate the effectiveness of our approach.
\vspace{-0.3cm}
\end{abstract}
\vspace{-0.3cm}
\begin{figure*}[htb!]
\includegraphics[width=\textwidth, page=1 ,trim={4.5cm 0cm 5.1cm 0cm},clip]{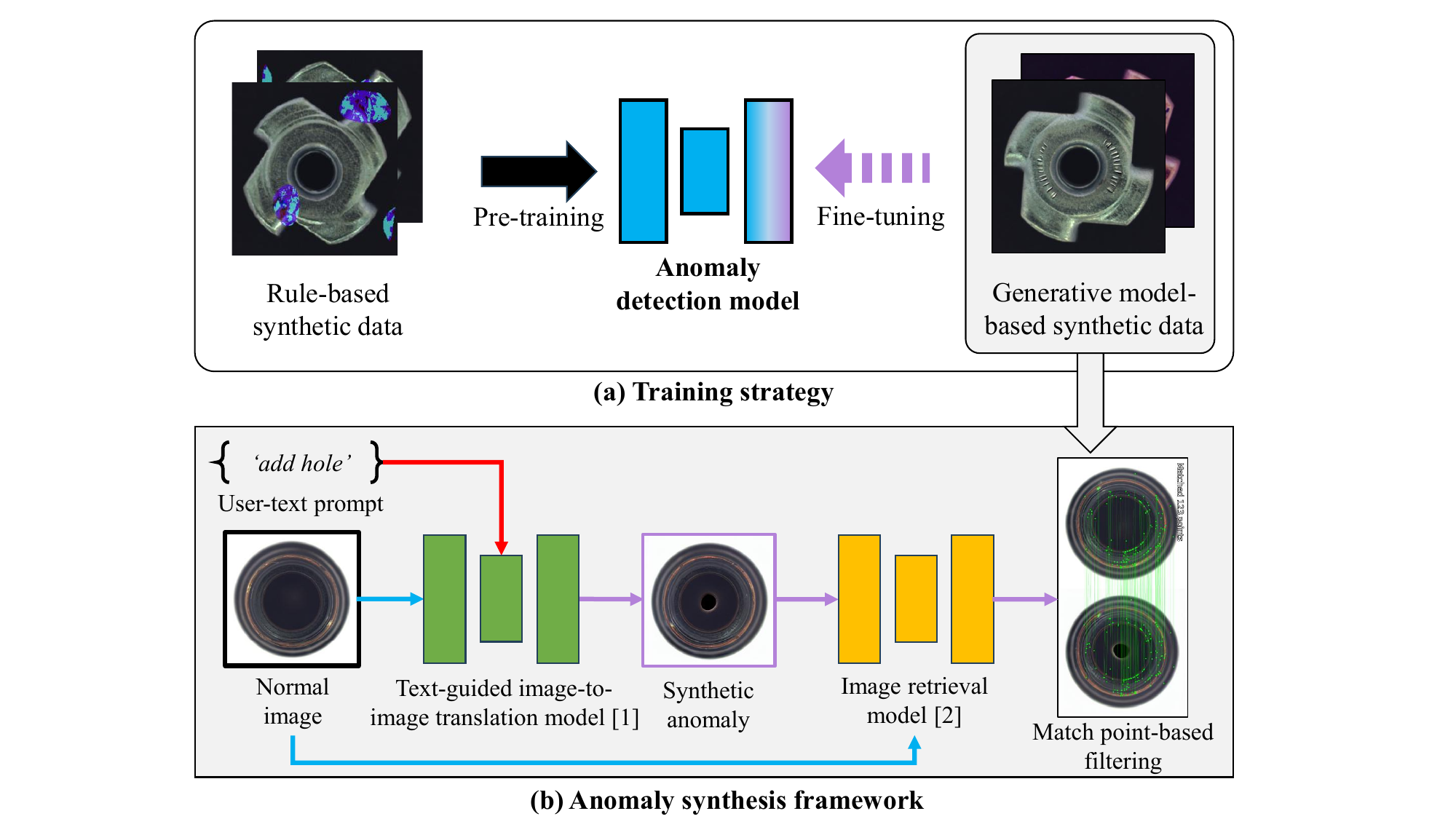} 
  \caption{Overview of the proposed approach. (a) To effectively leverage synthetic images, we introduce a novel training strategy that sequentially leverages large-scale rule-based synthesis for pre-training, followed by small-scale generative model-based synthesis for fine-tuning. (b) To provide high quality data for the fine-tuning stage, we also propose an effective anomaly synthesis framework. It consists of a pre-trained text-guided image-to-image translation model \cite{magicbrush} and a pre-trained image retrieval model \cite{lightglue}. Given a normal image and a prompt that describes defect, the pre-trained text-guided image-to-image translation model synthesizes a synthetic anomaly. Since the text-guided image-to-image translation model in this work is pre-trained on MS COCO dataset \cite{coco}, it occasionally generates irrelevant results, we additionally employ a filtering mechanism. The image retrieval model compares the normal image with the synthetic anomaly, filtering out irrelevant synthetic anomalies based on the number of matched points.}
\end{figure*}
\vspace{-0.3cm}
\section{Introduction}
\label{introduction}
Detection of defective products early is critical to maximize productivity in industrial manufacturing. On the other hand, obtaining real defect images in production environments is often difficult, making it challenging to construct paired data for supervised learning. Due to this limitation, the majority of prior studies have focused on unsupervised approaches that only utilize normal samples.
To overcome this data scarcity, various synthetic defect image generation strategies have been proposed. These strategies are broadly categorized into rule-based synthesis and generative model-based synthesis.

Rule-based synthesis involves generating defects by directly manipulating normal images through simple transformations such as Cut-Paste \cite{cutpaste} or injecting noise \cite{draem}. These methods are cost-effective and offer explicit control over defect attributes like size, shape, and position. However, they frequently fail to reproduce the complexity of real-world defects, resulting in synthetic images that often lack realism. Consequently, relying on such low-fidelity data for training negatively impacts the reliability of downstream models, ultimately yielding in reduced detection accuracy.

In contrast, generative model-based synthesis, which leverages generative models like GANs (e. g., CycleGAN \cite{cyclegan} and StyleGAN \cite{stylegan}) or Diffusion Models (e. g., stable diffusion \cite{stablediffusion}), is capable of producing synthetic defect images with high fidelity and diversity. Despite these advantages, generative model-based approaches are inherently resource-intensive that require substantial computational power and time for both the training and sampling phases. Furthermore, their high data dependency and slow inference speeds pose significant hurdles to practical deployment.

Motivated by these challenges, we propose a novel training-free defect image synthesis framework as illustrated in Fig. 1. The proposed framework utilizes a pre-trained text-guided image-to-image translation model to convert a normal sample into a defective image by taking both the normal sample and a text prompt describing the defect as inputs. However, since pre-trained text-guided image-to-image translation models are typically trained on large-scale datasets unrelated to specific industrial defects, they may generate irrelevant or unwanted results as shown in Fig. 2. To address this, we additionally employ a filtering mechanism utilizing a pre-trained image retrieval model. By comparing the generated synthetic anomaly with the normal image, the model automatically filters out irrelevant results. Consequently, our proposed method enables the generation of high-quality defect images without the need for additional training.

Regarding the utilization of synthetic data, prior studies have explored various training strategies, broadly categorized into those relying solely on synthetic images \cite{popp1} and those leveraging both synthetic and real images \cite{fan,kazdan,popp2,Qraitem,Kirichenko}. Given the data scarcity in anomaly detection, training needs to primarily rely on synthetic data, but generating the required large volume of high-fidelity data via generative models incurs high computational costs.

To address this challenge, we propose a two-stage training strategy that leverages both rule-based and generative model-based synthesis. In the first stage, we establish a strong baseline by pre-training the model on a large scale using low-cost rule-based synthetic data. Subsequently, we fine-tune this pre-trained model using a small set of high-quality samples generated by our framework. This approach drastically reduces the required number of computationally expensive generative model-based samples, thereby significantly minimizing the overall sampling cost.

To the best of our knowledge, this is the first work to propose a training-free defect synthesis framework that incorporates a retrieval-based filtering mechanism to ensure structural consistency. Furthermore, we are the first to introduce a cost-aware hybrid training strategy that sequentially utilizes rule-based and generative synthesis to optimize the trade-off between the data generation cost and the detection performance. We validate the efficacy of the proposed approach through extensive experiments on the MVTec AD dataset \cite{mvtec}.

In summary, our key contributions are:
\vspace{-0.25cm}
\begin{itemize}
    \item We present a novel training-free defect image synthesis framework that leverages a pre-trained text-guided image-to-image translation model and an image retrieval model to generate realistic defect images. The generated defect images are provided for fine-tuning of the model.

    \item We introduce a cost-efficient two-stage training strategy that sequentially leverages large-scale rule-based synthesis for pre-training and small-scale generative model-based synthesis for fine-tuning. This method effectively serves as an inexpensive solution to the data scarcity problem.
\end{itemize}

\section{Related works}
\label{related work}
\subsection{Synthetic defect image generation}
\label{synthetic defect}
Industrial anomaly detection is a critical component of modern manufacturing systems. In practice, it is often difficult to collect sufficient defective samples, leading most prior works to focus on unsupervised approaches that rely solely on normal data. Classical supervised approaches generally remove a region of an image and attempt to inpaint the missing part to infer potential defects. Recently, with the rapid advancement of generative models, synthesizing defective images for supervised training has gained substantial attention. Existing synthetic defect generation methods can be categorized into two groups: rule-based synthesis and model-based synthesis.

Rule-based methods generate defects by directly manipulating normal images through simple transformations. Cut-Paste \cite{cutpaste} removes image patches from one normal sample and pastes them onto another. ADSAS \cite{adsas} extends this idea by applying color transformations, blurring, and noise injection to the extracted patches before compositing. NSA \cite{nsa} removes a region from a normal image and synthesizes defects using Poisson harmonization to achieve more natural blending. CDO \cite{cdo} and RD++ \cite{rd} create abnormal samples by injecting Gaussian noise into normal images. Other approaches introduce patches from the DTD \cite{dtd} texture dataset into normal images, while varying the mask shapes and placements through Perlin noise, as demonstrated in DRAEM \cite{draem} and DESTSEG \cite{detseg}, whereas MEMSEG \cite{memseg} restricts anomaly injection to foreground regions. These algorithmic approaches offer explicit control over defect size, shape, and position with low computational cost, yet they often fail to reproduce the complexity of real world defects, limiting realism and diversity.

In contrast, generative-model–based synthesis approaches have emerged as powerful alternatives due to their ability to produce highly realistic defects. Early studies leveraged GANs architectures such as CycleGAN \cite{cyclegan}, StyleGAN \cite{stylegan}, and conditional GANs. SDGAN \cite{sdgan}, AttenCGAN \cite{attngan}, and Defect-GAN \cite{defectgan} adopt CycleGAN to learn unpaired mappings between normal and defective domains. DFMGAN \cite{dfmgan} exploits the expressive latent space of StyleGAN to manipulate defect-related latent factors. Conditional GAN variants, such as those used in MDGAN \cite{mdgan} and DCDGAN \cite{dcdgan}, provide explicit control over defect attributes including type, geometry, and spatial location.

With diffusion models surpassing GANs in fidelity, several works began training DDPM \cite{ddpm} from scratch. For example, RealNet \cite{realnet} incorporates a strength-controllable perturbation term into the reverse diffusion process after training on normal-only data. Although these approaches demonstrate strong generative capability, their substantial training cost, large data requirements, and limited cross-domain generalization hinder practical deployment. To reduce the computational burden, subsequent research explored leveraging pre-trained diffusion models (e.g., Stable Diffusion \cite{stablediffusion}) or fine-tuning them for defect synthesis. AnomalyDiffusion \cite{anomalydiffusion} disentangles appearance and localization factors to generate diverse anomalies. DualAnoDiff \cite{dualanodiff} employs dual diffusion processes to jointly generate an anomalous sample and its corresponding defect region. AdaBLDM \cite{adabldm} introduces adaptive latent blending with tri-map masks for localized anomaly insertion. CAGen \cite{cagen} uses a ControlNet-augmented Stable Diffusion pipeline fine-tuned with binary masks and text prompts, enabling explicit control over defect placement and style. Other methods, such as SeaS \cite{seas}, incorporate separation-and-sharing strategies within the U-Net to model both normal and abnormal variations, while AnomalyXFusion \cite{anomalyxfusion} extends the conditioning space to multi-modal inputs by fusing images, masks, and textual features. Although these methods dramatically reduce training costs compared to training diffusion models from scratch, they still require fine-tuning procedures.

Most recently, training-free approaches have gained attention. AnomalyAny \cite{anomalyany} synthesizes defects directly on top of Stable Diffusion using attention-guided optimization and prompt-driven refinement, enabling realistic and diverse anomalies without any additional training. However, due to its training-free nature, AnomalyAny is highly sensitive to the interplay between the input normal image and the textual prompt, often producing inconsistent results—particularly problematic when generating large-scale augmented datasets. Such variability can degrade downstream model performance, necessitating quality-control strategies to filter unstable or low-fidelity synthetic samples. To address this limitation, we incorporate an image retrieval–based filtering mechanism to automatically identify and discard incorrectly synthesized images, improving overall dataset quality.

\begin{figure*}[ht]
\includegraphics[width=\textwidth, page=2 ,trim={1.7cm 0.1cm 1.7cm 0cm},clip]{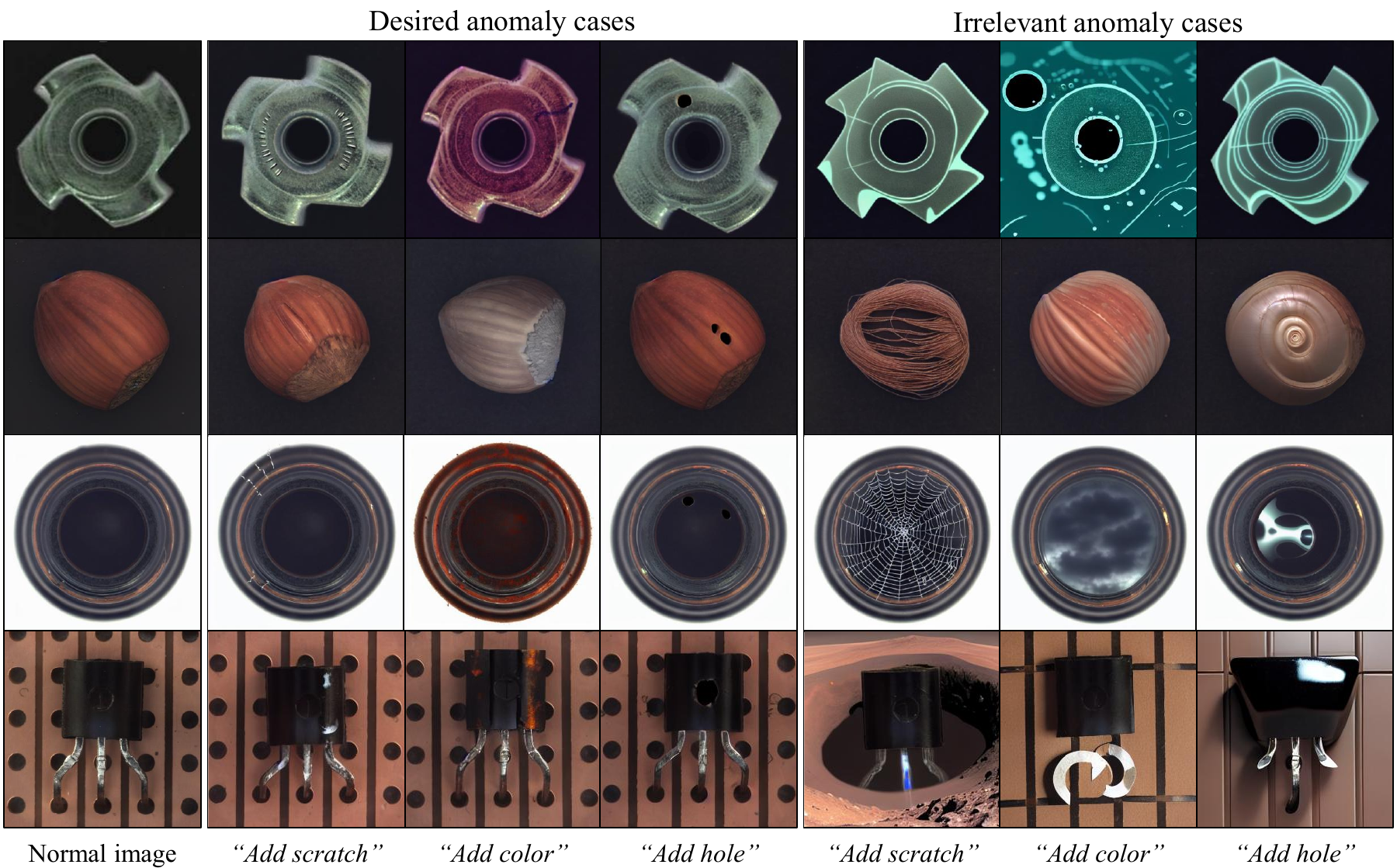} 
  \caption{Examples of synthetic defect images that are translated from a normal image. The `Desired anomaly cases' demonstrate high-fidelity synthesis where the generated defects are seamlessly integrated into the object surface while preserving the structural integrity and background context of the original normal image. In contrast, the `Irrelevant anomaly cases' exhibit significant failure modes, such as severe background distortions, structural inconsistencies, and the introduction of irrelevant artifacts that deviate from realistic defect characteristics. While integrating high-quality synthetic samples into the training pipeline is crucial for enhancing the anomaly detection performance in data-scarce regimes, training on inconsistent data disrupts the learning of discriminative features, leading to model degradation. These observations underscore the critical necessity of a filtering mechanism to  utilize high-fidelity synthesis.} 
\end{figure*}

\subsection{Training strategy using synthetic data}
\label{training strategy}
With the improvement in the quality of synthetic images, many studies have explored strategies for utilizing synthetic images as training data. Fan \textit{et al}. \cite{fan} demonstrate that while increasing the volume of synthetic data for training improves performance, it remains less efficient than real data due to limited diversity, though highly effective in data scarcity problems. To bridge the domain gap and improve adaptation, Qraitem \textit{et al}. \cite{Qraitem} proposed a novel framework involving pre-training on synthetic data and fine-tuning on real data. Kirichenko \textit{et al}. \cite{Kirichenko} introduced deep feature reweighting to retrain only the final layer, minimizing reliance on background spurious correlations. Addressing data quality and robustness, Kazdan \textit{et al}. \cite{kazdan} utilized clustering-based filtering to prevent model collapse by selecting synthetic samples resembling real distributions. Popp \textit{et al}. \cite{popp1} applied teacher-student distillation to mitigate spurious features arising from synthetic data’s lack of diversity. 

Since acquiring real defect images for anomaly detection is extremely challenging, it would be best if we could rely on synthetic data for training. However, model training requires a large volume of synthetic defect images, which leads to high computational costs. To address this problem, we propose an efficient training strategy that leverages both rule-based synthesis and generative model-based synthesis. This method significantly reduces the computational burden of data generation.

\begin{figure*}[t!]
\includegraphics[width=\textwidth, page=3 ,trim={3.9cm 0cm 3.9cm 0cm},clip]{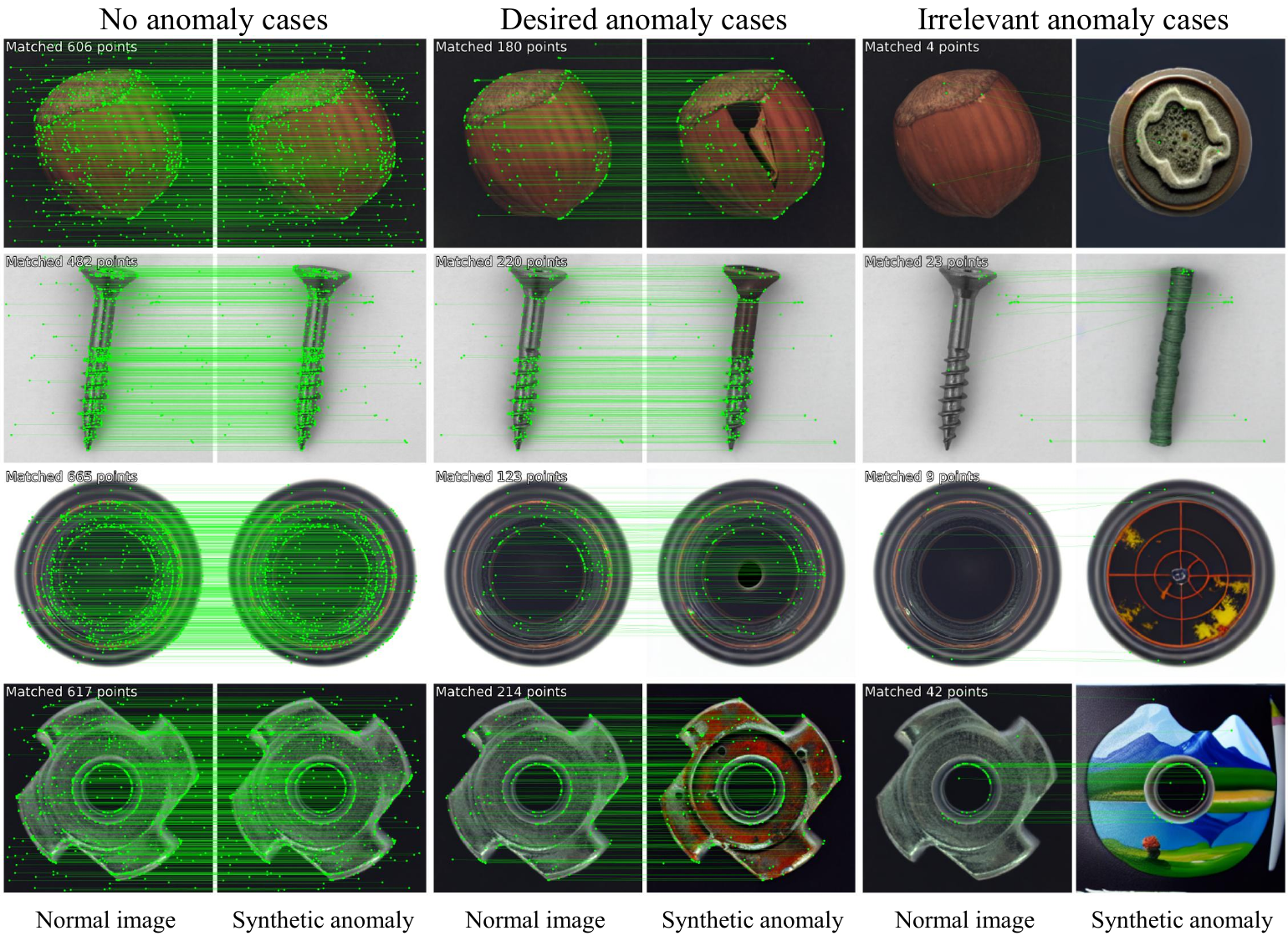} 
  \caption{Visualization of the filtering mechanism based on feature matching. We compare the normal image with the synthetic anomaly. The underlying hypothesis is that a well-generated defect image should retain high structural correspondence with the original image, except for the localized defect region. The green lines represent the matching points between the two images. In the `No anomaly cases', the image remains almost identical to the original, resulting in the highest number of matching points. In the `Desired anomaly cases', fewer points are matched than in the `No anomaly cases' due to the generated defects. However, there are still many matching points, meaning the structure information is well preserved. Finally, in the `Irrelevant anomaly cases', there are very few matching points. This indicates that we can filter out meaningless synthetic defect images based on the number of matching points.}
\end{figure*}

\section{Method}
\label{method}
\subsection{Synthetic defect image generation with pre-trained models}
\label{synthetic defect using pre-trained model}
As illustrated in Fig. 1, our anomaly synthesis framework consists of two core components: a text-guided image-to-image translation module for synthetic defect generation and a deep feature matching module for quality verification. Recently, numerous text-guided diffusion models have been introduced to modify images based on natural language prompts \cite{instructpix2pix,magicbrush}. However, \cite{instructpix2pix} often performs global transformations, altering the entire image structure rather than focusing on specific regions. This global change is unsuitable for anomaly detection tasks, where preserving the structural integrity of the non-defective background is paramount. 

In contrast, local editing model \cite{magicbrush} is designed to enable precise, local editing of specific image regions while maintaining the original context. Therefore, we employ a pre-trained model as our generator to synthesize realistic defect images by providing prompts that explicitly describe the characteristics of potential abnormal patterns (e. g., scratch, dent, or stain). Despite its editing capabilities, leveraging pre-trained model in an industrial setting presents a significant challenge. Since the model is pre-trained on the MS COCO dataset \cite{coco}, which primarily consists of natural scenes, there exists a domain gap when applied to industrial objects. Consequently, the model occasionally suffers from semantic drift or structural inconsistency, generating results that are completely unrelated to the input image as demonstrated in the `Irrelevant anomaly cases' of Fig. 2. Directly utilizing such inconsistent samples for training, confusing the anomaly detection model and severely degrading its performance. 

To mitigate this issue and ensure the quality of the training data, we employ filtering mechanism that utilizes image retrieval model \cite{lightglue} for local feature matching. Image retrieval model matches local keypoints between the normal image and the synthesized defect image to evaluate their structural similarity, as shown in Fig. 3. The underlying hypothesis is that a well-generated defect image should retain high structural correspondence with the original image, except for the localized defect region. Filtering mechanism effectively filters out irrelevant images—specifically those with excessive structural changes (too few matches) or failed edits (too many matches implying identity). This filtering process ensures that only high-fidelity synthetic defects that preserve the object's geometry are utilized for training.

\subsection{Training strategy}
Exclusively relying on high-fidelity synthetic images generated by diffusion models is computationally infeasible, as it requires massive sampling to cover the diverse defects. To overcome this limitation, we propose a two-stage training framework that leverages both rule-based synthesis and generative model-based synthesis. In the first stage, we utilize  rule-based synthesis techniques to generate a massive volume of defect samples with low cost.
This phase constructs a solid foundation for the reconstruction task, enabling the network to effectively restore the normal appearance from inputs corrupted by low-level structural anomalies. This establishes a strong baseline performance without the computational overhead of iterative sampling.
In the second stage, we fine-tune the model using a smaller curated set of high-fidelity images generated by our framework. While the pre-training phase constructs a solid foundation for basic restoration, this fine-tuning step is essential for bridging the domain gap between simply synthesized defects that cannot occur in real industrial manufacturing environments and realistic industrial defects. This ensures the model aligns its reconstruction capability with real-world distributions, achieving high performance with minimal computational overhead.

\begin{figure*}[t!]
\includegraphics[width=\textwidth, page=4 ,trim={8.4cm -0.5cm 8.2cm -0.5cm},clip]{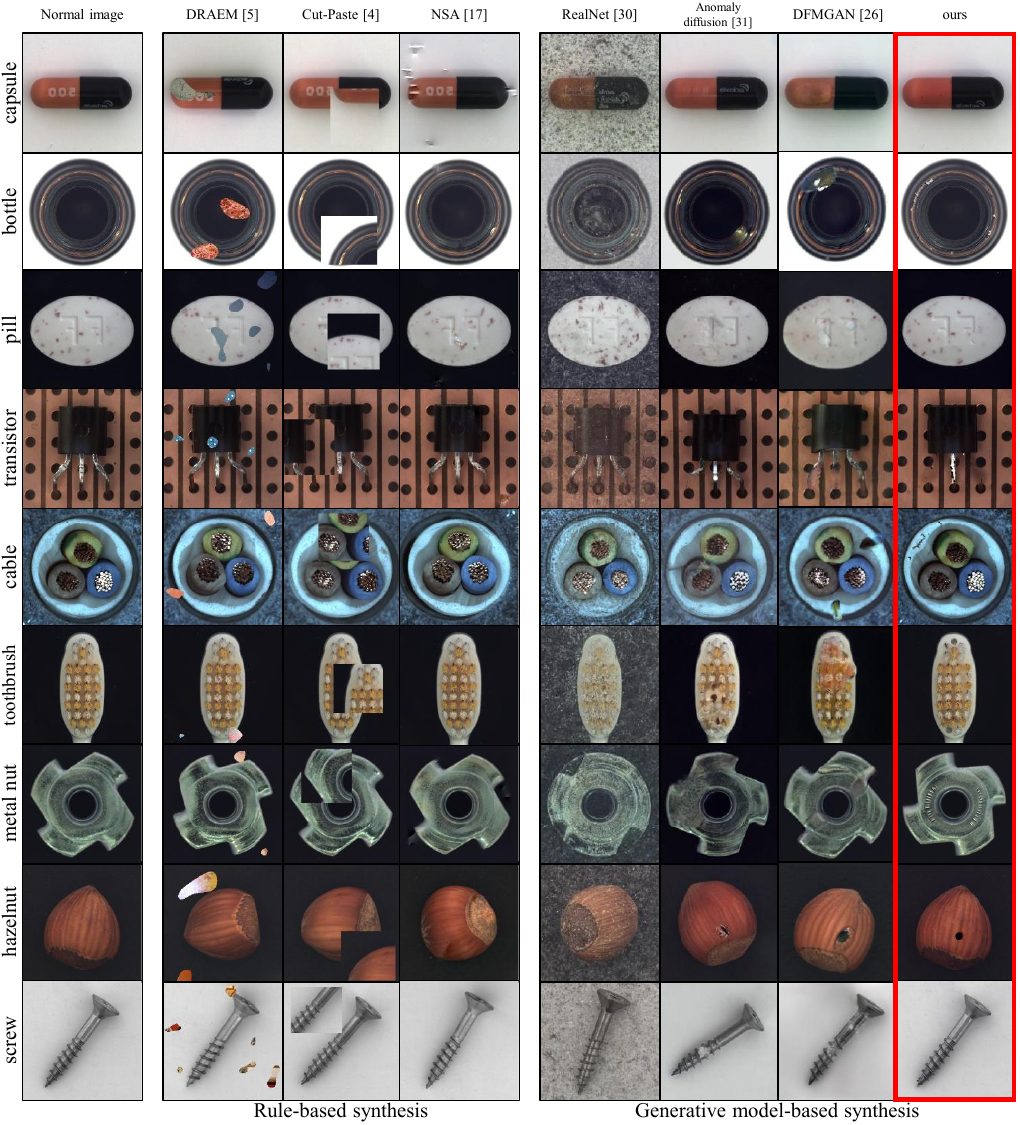} 
  \caption{Qualitative comparison of defect image synthesis methods across various object categories from the MVTec AD dataset \cite{mvtec}. From left to right: DRAEM \cite{draem}, Cut-Paste \cite{cutpaste}, NSA \cite{nsa}, RealNet \cite{realnet}, AnomalyDiffusion \cite{anomalydiffusion}, DFMGAN \cite{dfmgan}, and ours. Defect images synthesized using rule-based synthesis are often unrealistic and do not resemble defects that can occur in real-world industrial manufacturing environments. In contrast, generative model-based synthesis approaches generally produce more realistic defect images. Our proposed method, despite relying solely on pre-trained models, is capable of generating defect images that are even more realistic than those produced by existing generative synthesis methods.}
\end{figure*}

\section{Experiments}
\label{experiments}
For experiments, we have employed the publicly available anomaly detection dataset, MVTec AD \cite{mvtec}. This dataset consists of nine object categories and six texture categories. Among these, we have only used nine object categories such as capsule, bottle, pill, transistor, cable, toothbrush, metalnut, hazelnut, and screw for our experiments. 
For the simple synthesis, we adopted the anomaly generation module proposed in DRAEM \cite{draem}. This method simulates structural anomalies by blending external textures (e. g., from DTD \cite{dtd}) or augmented source images with the original normal images using Perlin noise masks
For generative model-based synthesis, three hundred synthetic defect images are generated using our framework.
We employ the pre-trained text-guided image-to-image translation model \cite{magicbrush} and the pre-trained image retrieval model \cite{lightglue} that are available from their official GitHub repositories.

\subsection{Qualitative results}
\label{Qualitative}
To comprehensively evaluate the perceptual quality and realism of the generated samples, we present a qualitative comparison between our proposed framework and existing synthesis methods in Fig. 4.
The baseline methods are categorized into two groups: rule-based synthesis (DRAEM \cite{draem}, Cut-Paste \cite{cutpaste}, NSA \cite{nsa}) and generative model-based synthesis (RealNet \cite{realnet}, AnomalyDiffusion \cite{anomalydiffusion}, DFMGAN \cite{dfmgan}).

As observed in Fig. 4, rule-based synthesis methods exhibit significant limitations in mimicking real-world industrial defects. Cut-Paste simply overlays patches without any blending mechanism, producing unnatural boundary discontinuities that look artificial. NSA employs Poisson blending to soften edges but often introduces interpolation artifacts. DRAEM, relying on stochastic Perlin noise and external texture dataset, yields incoherent artifacts that do not align with the patterns that observed in industrial defects. Such unrealistic samples may prevent the anomaly detection model from learning the subtle characteristics of actual defects, leading to suboptimal performance.

In contrast, existing generative model-based approaches, RealNet, AnomalyDiffusion, and DFMGAN in Fig. 4 demonstrate improved visual fidelity compared to rule-based synthesis. Although generative baselines outperform upon rule-based methods, they exhibit distinct limitations. RealNet tends to introduce excessive global noise across the entire image, degrading the visual quality of the non-defective background. Meanwhile, AnomalyDiffusion and DFMGAN generally produce realistic results but struggle with complex categories; specifically, they fail to generate plausible defects for the toothbrush and capsule classes.

Our proposed method consistently generates high-fidelity synthetic defects across all categories, including the challenging toothbrush and capsule classes. By effectively leveraging text-guided image-to-image translation model with image retrieval model, our framework successfully synthesizes realistic anomalies that preserve the integrity of the original object, outperforming both rule-based and generative baselines in terms of visual realism and diversity.

\begin{figure*}[t!]
\includegraphics[width=\textwidth, page=5 ,trim={4.2cm 6cm 3cm 5cm},clip]{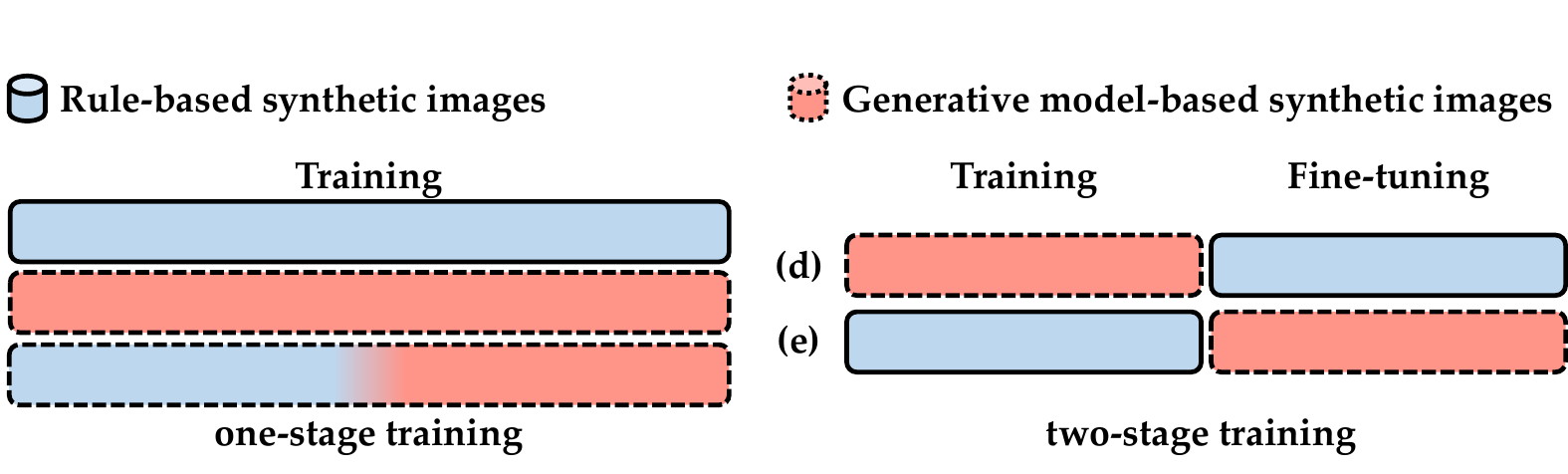} 
  \caption{Overview of the training pipeline. Blue boxes represent defect images generated by rule-based synthesis, while red boxes represent generative model-based synthetic images. Strategies (a), (b), and (c) use only one-stage training, whereas strategies (d) and (e) employ a two-stage training process with fine-tuning.}
\end{figure*}

\begin{table}[t!]
\caption{Quantitative comparison of anomaly detection performance (AUROC) across different training strategies. We evaluate the effectiveness of utilizing rule-based synthetic images and generative model-based images under various configurations. Strategies (a)-(c) represent single-stage training, while (d) and ours involve a two-stage process. While the large-scale sim dataset establishes a generalized baseline (a), utilizing gen data alone (b) or simply mixing datasets (c) yields suboptimal results due to data scarcity or the dominance of rule-based synthetic samples. Furthermore, fine-tuning in the reverse order (d) leads to a significant performance drop (77.4$\%$) attributed to catastrophic forgetting. In contrast, our strategy achieves the best performance with an average of 98.7$\%$, demonstrating that pre-training on large-scale  rule-based synthetic data followed by fine-tuning on high-fidelity generative images enables the model to effectively adapt to realistic anomalies.}
\resizebox{\textwidth}{!}{
\begin{tabular}{lccccc}
& \multicolumn{3}{c}{{$\longleftarrow$} \quad\quad\quad\quad \textit{one-stage training } \quad \quad \quad\quad {$\longrightarrow$}}  & \multicolumn{2}{c}{{$\longleftarrow$}\quad \textit{two-stage training } \quad{$\longrightarrow$}}             \\ \hline
\multicolumn{1}{l|}{\diagbox{class}{strategy}} & \multicolumn{1}{c|}{\quad\quad\: (a) \:\quad\quad\quad}            & \multicolumn{1}{c|}{\quad\quad\: (b) \:\quad\quad\quad}    & \multicolumn{1}{c|}{\quad\quad\: (c) \:\quad\quad\quad}            & \multicolumn{1}{c|}{\quad\quad\: (d) \:\quad\quad\quad}    & \quad ours\quad\quad \\ \hline\hline
\multicolumn{1}{l|}{capsule}                                      & \multicolumn{1}{c|}{97.3}         & \multicolumn{1}{c|}{87.1} & \multicolumn{1}{c|}{97.4}         & \multicolumn{1}{c|}{54.2} & \textbf{98.7} \\ \hline
\multicolumn{1}{l|}{bottle}                                       & \multicolumn{1}{c|}{96.7}         & \multicolumn{1}{c|}{94.8} & \multicolumn{1}{c|}{96.3}         & \multicolumn{1}{c|}{96.1} & \textbf{99.0} \\ \hline
\multicolumn{1}{l|}{pill}                                         & \multicolumn{1}{c|}{98.2}         & \multicolumn{1}{c|}{81.8} & \multicolumn{1}{c|}{97.0}         & \multicolumn{1}{c|}{76.2} & \textbf{98.4} \\ \hline
\multicolumn{1}{l|}{transistor}                                   & \multicolumn{1}{c|}{93.1}         & \multicolumn{1}{c|}{89.8} & \multicolumn{1}{c|}{92.1}         & \multicolumn{1}{c|}{83.1} & \textbf{95.9} \\ \hline
\multicolumn{1}{l|}{cable}                                        & \multicolumn{1}{c|}{97.8}         & \multicolumn{1}{c|}{70.4} & \multicolumn{1}{c|}{92.2}         & \multicolumn{1}{c|}{80.9} & \textbf{97.1} \\ \hline
\multicolumn{1}{l|}{toothbrush}                                   & \multicolumn{1}{c|}{99.7}         & \multicolumn{1}{c|}{99.4} & \multicolumn{1}{c|}{99.4}         & \multicolumn{1}{c|}{77.8} & \textbf{100}  \\ \hline
\multicolumn{1}{l|}{metal nut}                                    & \multicolumn{1}{c|}{98.7}         & \multicolumn{1}{c|}{84.4} & \multicolumn{1}{c|}{97.5}         & \multicolumn{1}{c|}{64.1} & \textbf{99.1} \\ \hline
\multicolumn{1}{l|}{hazelnut}                                     & \multicolumn{1}{c|}{\textbf{100}} & \multicolumn{1}{c|}{88.0} & \multicolumn{1}{c|}{\textbf{100}} & \multicolumn{1}{c|}{94.3} & \textbf{100}  \\ \hline
\multicolumn{1}{l|}{screw}                                        & \multicolumn{1}{c|}{93.9}         & \multicolumn{1}{c|}{89.5} & \multicolumn{1}{c|}{96.2}         & \multicolumn{1}{c|}{70.3} & \textbf{99.5} \\ \hline \hline
\multicolumn{1}{l|}{avg}                                          & \multicolumn{1}{c|}{96.6}         & \multicolumn{1}{c|}{87.2} & \multicolumn{1}{c|}{96.5}         & \multicolumn{1}{c|}{77.4} & \textbf{98.7} \\ \hline
\end{tabular}}

\end{table}

\subsection{Training strategy}
\label{Training}
To validate the effectiveness of proposed training strategy, we compared several training configurations using rule-based synthetic images and generative model-based synthetic images as illustrated in Fig. 5. Training strategies (a), (b), and (c) involve a single-stage training process with different types of datasets, whereas strategies (d) and (e) perform an additional fine-tuning stage after the initial training.

In case of (a) in Table 1, where the model was trained only on rule-based synthetic images, it achieves the highest performance among the single-stage strategies with an average of 96.6\%. Although the images are synthetically simple, the large scale of the dataset allowed the model to learn generalized representations.
In contrast, case (b) that trained exclusively on the dataset generated by our method achieves a lower accuracy of 87.2\% when trained solely on the proposed synthetic data. This highlights that synthesis quality serves as a potential solution but implies that adequate data quantity is essential to address scarcity issues.

In case of (c) in Table 1, simply mixing rule-based synthetic images and dataset generated by our method showed no significant difference from (a), despite a marginal decrease in the performance. This indicates that the dominance of the large-scale rule-based synthetic images overshadows the fewer images generated by our proposed method, limiting their potential contribution.
In the case of fine-tuning on rule-based synthetic images after pre-training on images generated by proposed method, shown in Table 1 (d), showed in the worst performance. We attribute this decrease in the performance to a dual failure: suffering from catastrophic forgetting of representations, and the inability of the limited fine-tuning schedule to accommodate the massive scale of rule-based synthetic images.

However, when the model is initially trained on rule-based synthetic images and subsequently fine-tuned with the images generated by the proposed method, it achieves the best performance. These results experimentally prove that pre-training with a large amount of rule-based synthetic images helps establish a strong baseline performance, and further fine-tuning with generative model-based synthetic images enables the model to better adapt to realistic anomalies, significantly improving its detection performance.

\subsection{Discussion}
\label{Discussion}
Our method successfully balances between the computational cost and the detection performance. However, there are still areas for improvement.
First, relying only on text prompts makes it difficult to control the generation process. We found that even small changes in the text can produce unwanted results. Also, text alone cannot describe the exact size or location of a defect. To fix this, we plan to leverage prompt optimization and dynamic guidance Classifier Free Guidance \cite{cfg} methods to make the generation more stable.
Second, different products have different types of defects. Our experiments show that some objects might not need as much training with high-quality generated images as complex objects do.

In the future, we will develop an adaptive training strategy. We plan to adjust the ratio of rule-based synthetic data to generative data depending on how complex the defect is. This approach will make our training process even more efficient.

\vspace{-0.1cm}
\section{Conclusion}
\label{conclusion}
Recent advances in synthetic image generation have led to growing interest in utilizing synthetic images as training data for anomaly detection. However, their high computational cost poses a challenge for large-scale deployment. We propose a framework that utilizes pre-trained models to synthesize realistic synthetic defect images. In addition, we introduce an effective training strategy that leverages both rule-based synthesis and generative model-based synthesis. Our experimental results confirm the effectiveness and practicality of the proposed framework and training strategy. In future work, we plan to further explore synthesis strategies tailored to different object types and defect categories and investigate methods for controlling the size and location of defects when generating high-quality anomaly images without additional model training.

\vspace{-0.1cm}
\subsubsection{Acknowledgements} This study was supported in part by the National Research Foundation of Korea (NRF) under Grant 2020R1F1A1048438, the High-performance Computing (HPC) Support project funded by the Ministry of Science and ICT and National IT Industry Promotion Agency (NIPA), and BK21 FOUR Project. 

%
%
%
%

We provide supplementary materials that cannot be included in the manuscript due to space limitation. We offer extra qualitative results and matching results.

\section{Extra results}
\label{related work}
The results for each category are presented in Figs. 1 - 9. The categories are include bottle, cable, capsule, hazelnut, metal nut, pill, screw, toothbrush, and transistor. The synthesized defect images exhibit high visual fidelity to the real MVTec AD samples and effectively simulate potential industrial defects that were not originally collected. We also provide additional results to demonstrate the effectiveness of the matching point-based filtering mechanism. The results for bottle, cable, capsule, hazelnut, metal nut, pill, screw, toothbrush, and transistor are shown in Figs. 10 - 18. Each figure contains `no anomaly cases', `desired anomaly cases', and `irrelevant cases'.

\begin{figure*}[b]
  \includegraphics[width=\textwidth, page=1 ,trim={1cm 0cm 1cm 0cm},clip]{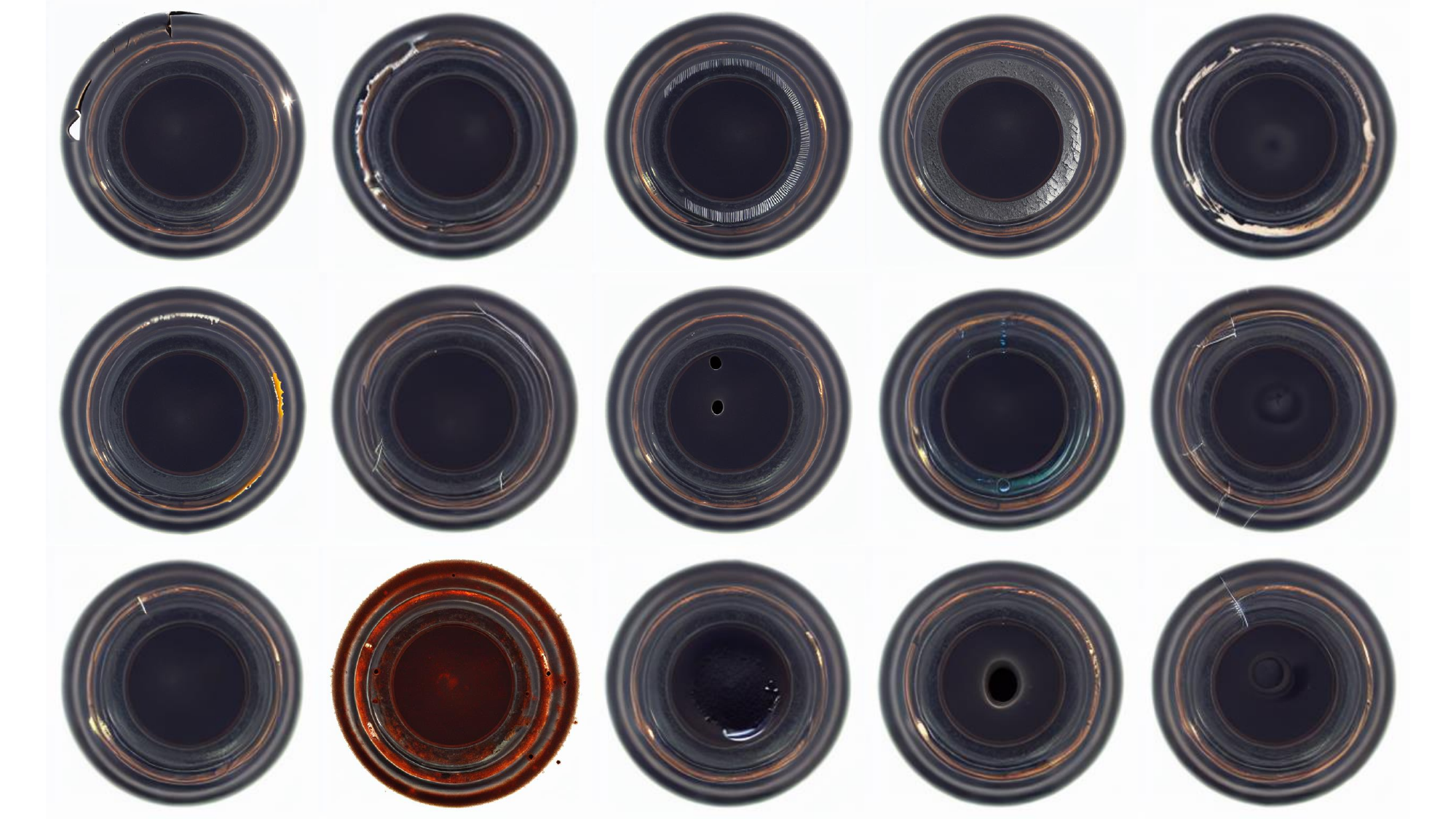} 
  \caption{Additional qualitative results: Synthetic bottle anomaly}
\end{figure*}

\vspace{3cm}

\begin{figure*}[t]
  \includegraphics[width=\textwidth, page=2 ,trim={1cm 0cm 1cm 0cm},clip]{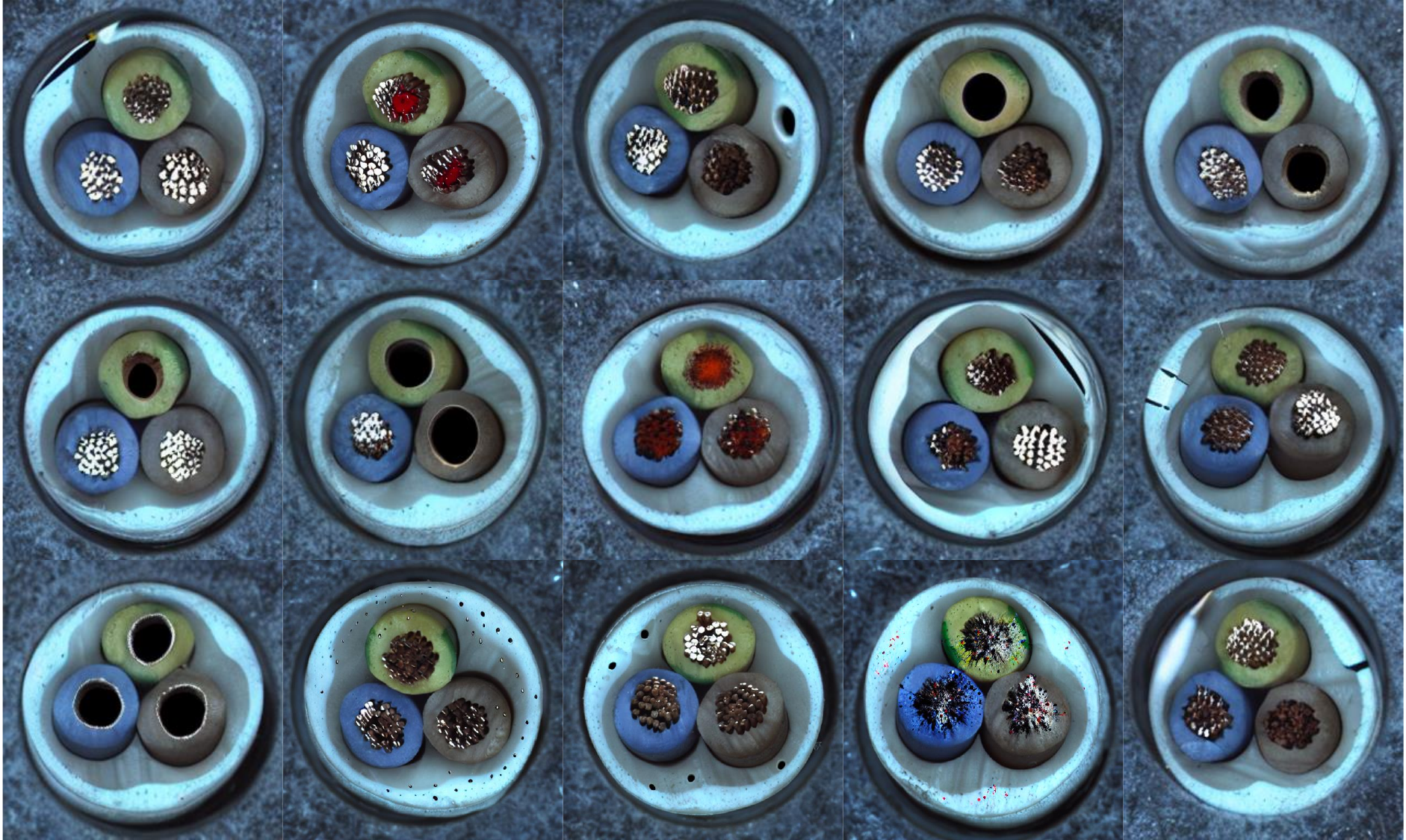} 
  \caption{Additional qualitative results: Synthetic cable anomaly}
\end{figure*}

\vspace{5cm}

\begin{figure*}[b]
    \includegraphics[width=\textwidth, page=3 ,trim={1cm 0cm 1cm 0cm},clip]{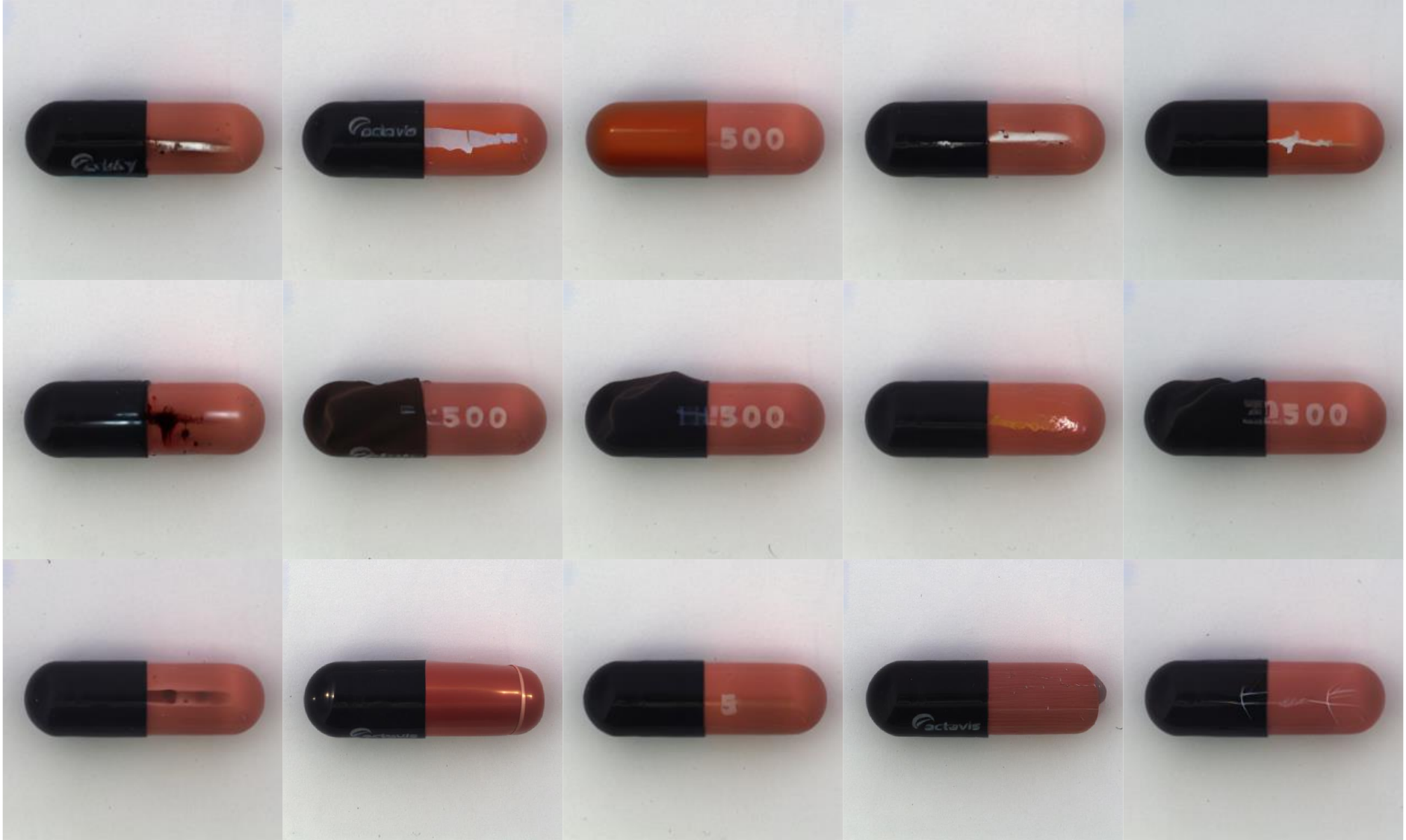} 
  \caption{Additional qualitative results: Synthetic capsule anomaly}
\end{figure*}

\begin{figure*}[t]
    \includegraphics[width=\textwidth, page=4 ,trim={1cm 0cm 1cm 0cm},clip]{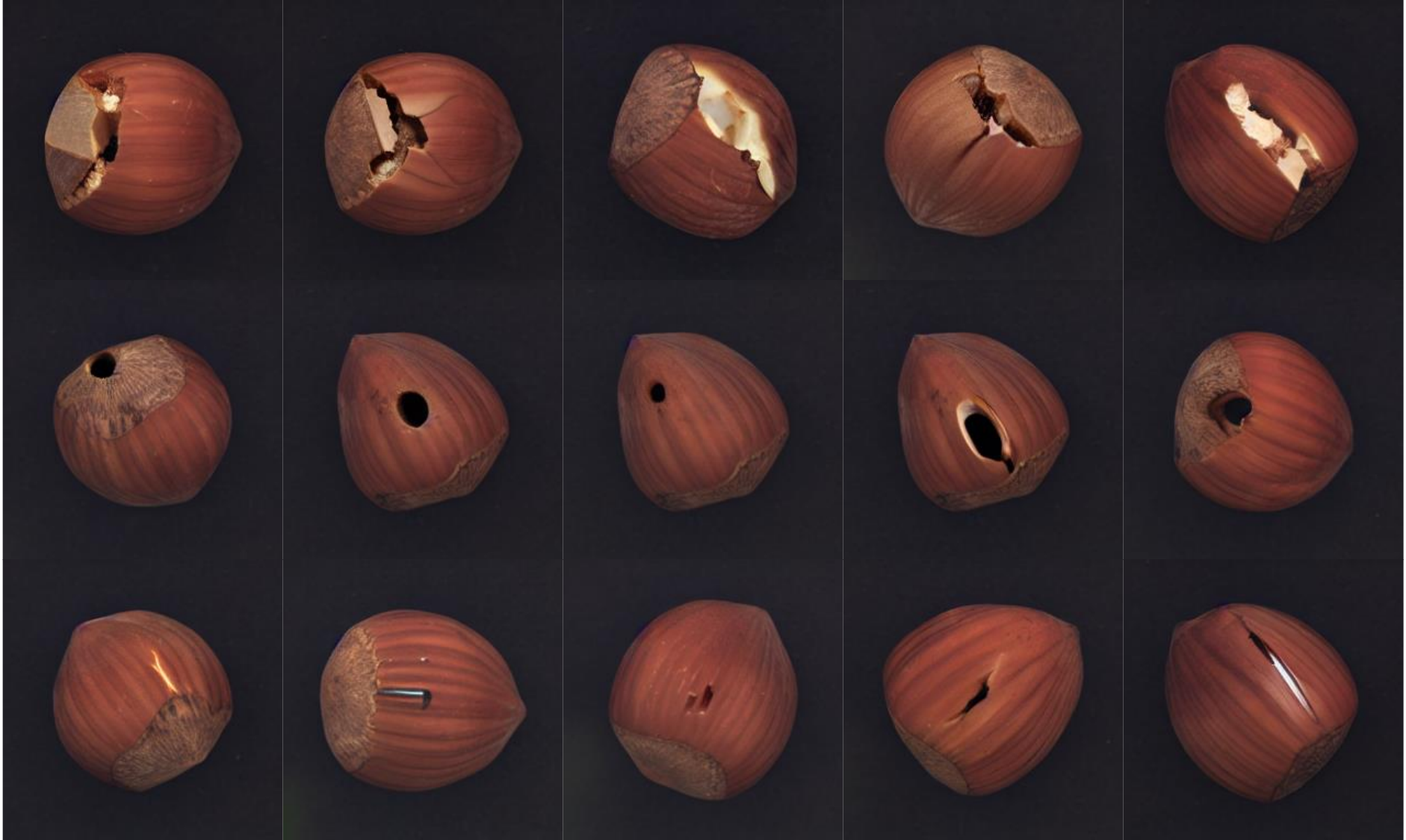} 
  \caption{Additional qualitative results: Synthetic hazelnut anomaly}
\end{figure*}

\vspace{5cm}

\begin{figure*}[b]
    \includegraphics[width=\textwidth, page=5 ,trim={1cm 0cm 1cm 0cm},clip]{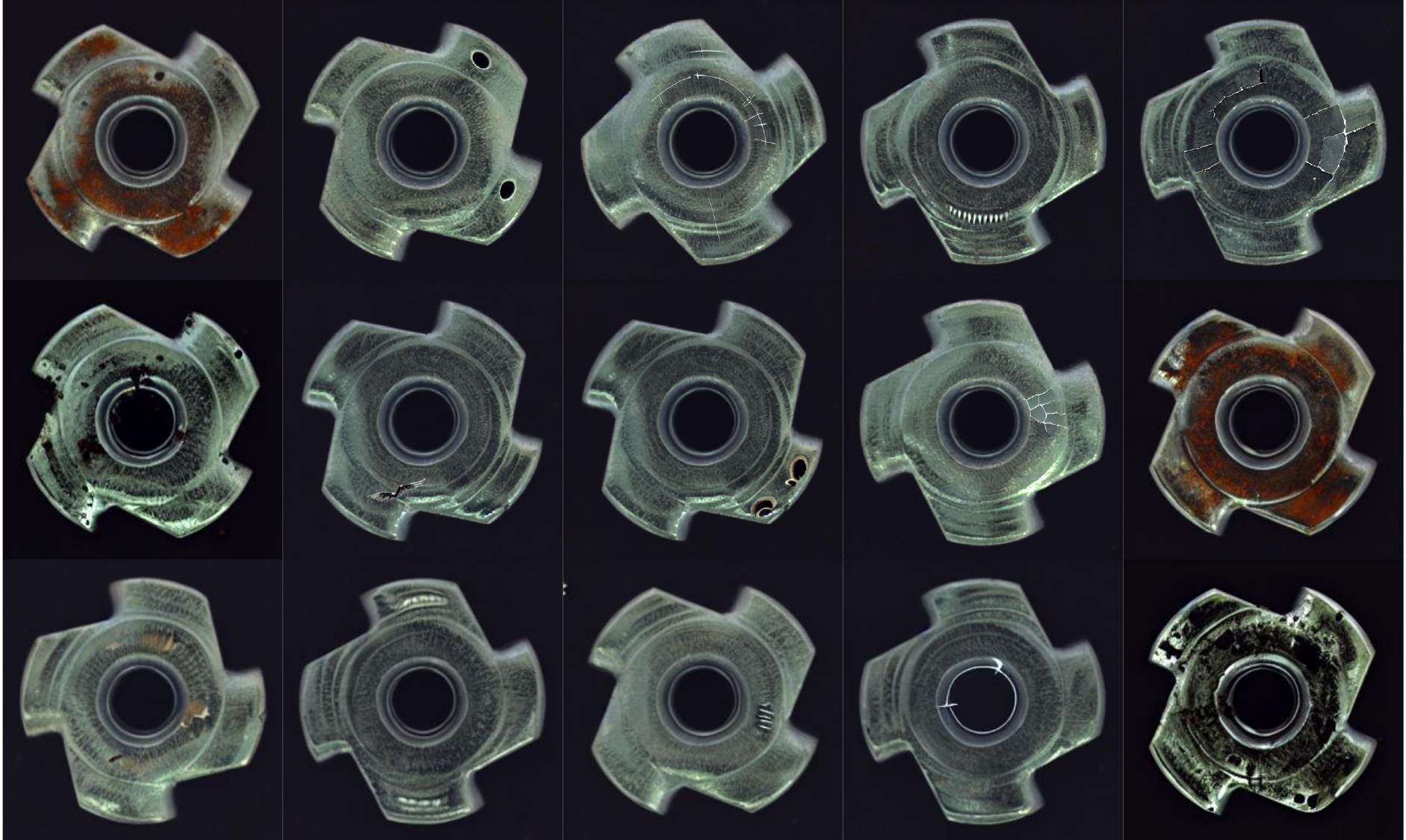} 
  \caption{Additional qualitative results: Synthetic metal nut anomaly}
\end{figure*}

\begin{figure*}[t]
    \includegraphics[width=\textwidth, page=6 ,trim={1cm 0cm 1cm 0cm},clip]{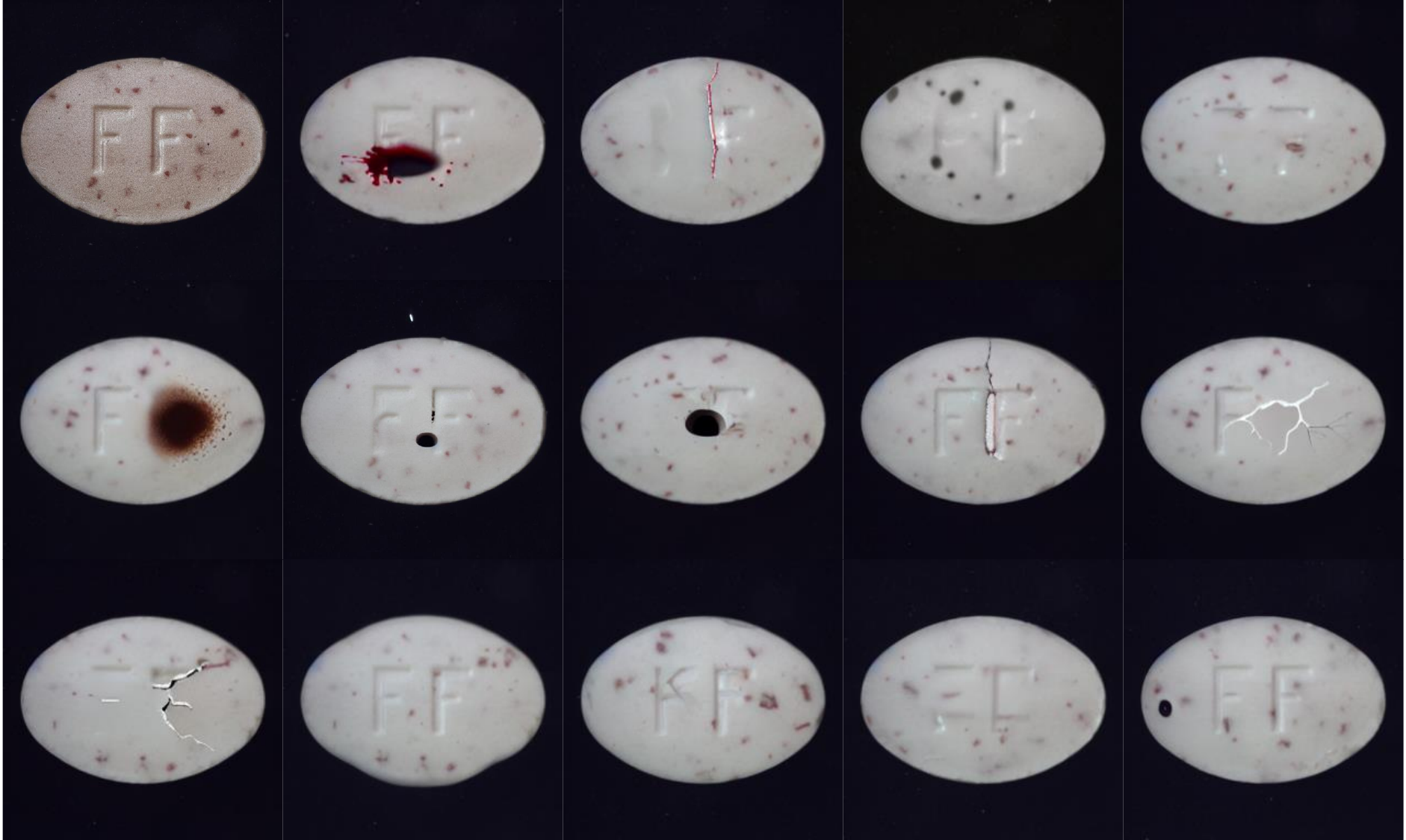} 
  \caption{Additional qualitative results: Synthetic pill anomaly}
\end{figure*}

\vspace{5cm}

\begin{figure*}[b]
    \includegraphics[width=\textwidth, page=7 ,trim={1cm 0cm 1cm 0cm},clip]{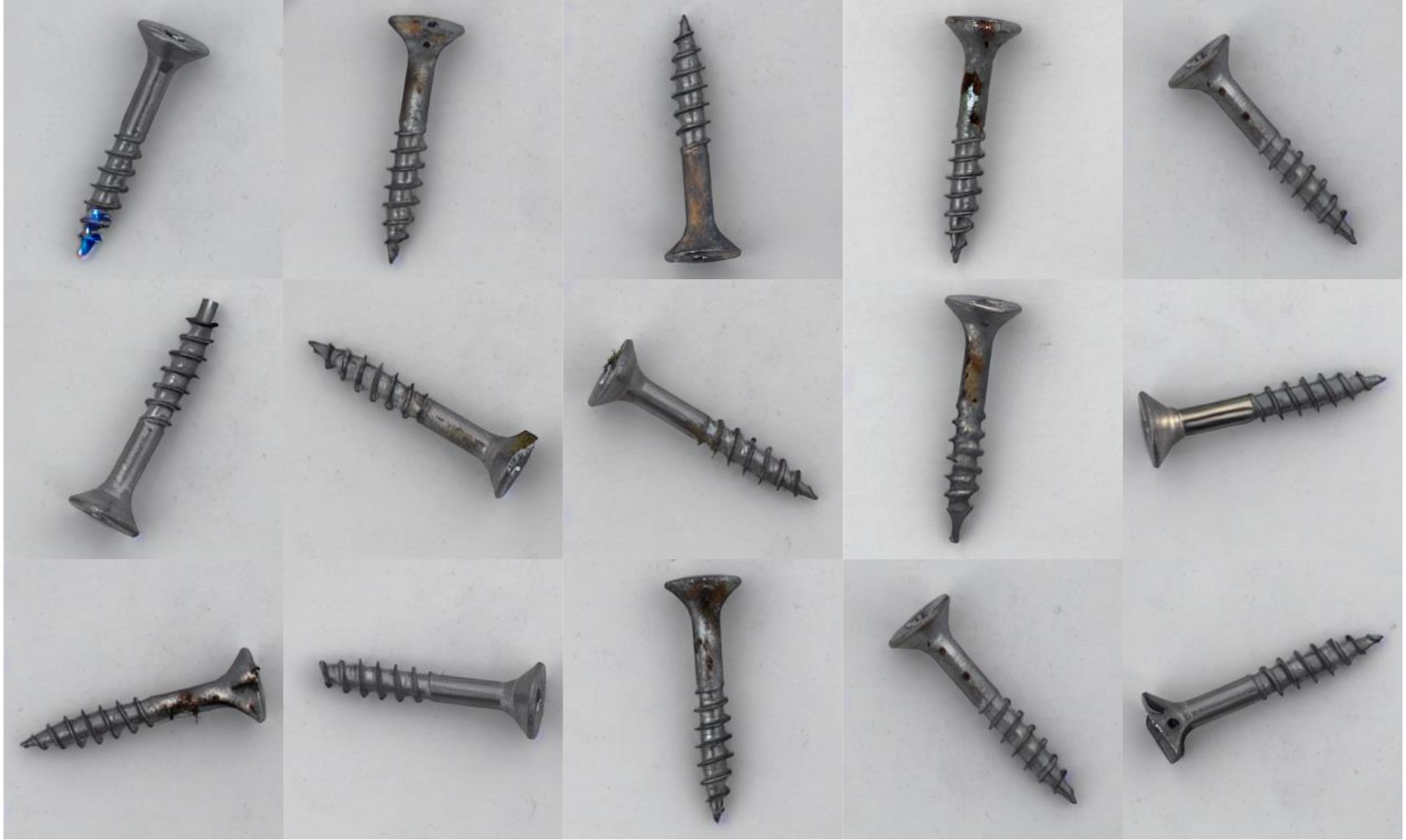} 
  \caption{Additional qualitative results: Synthetic screw anomaly}
\end{figure*}

\begin{figure*}[t]
    \includegraphics[width=\textwidth, page=8 ,trim={1cm 0cm 1cm 0cm},clip]{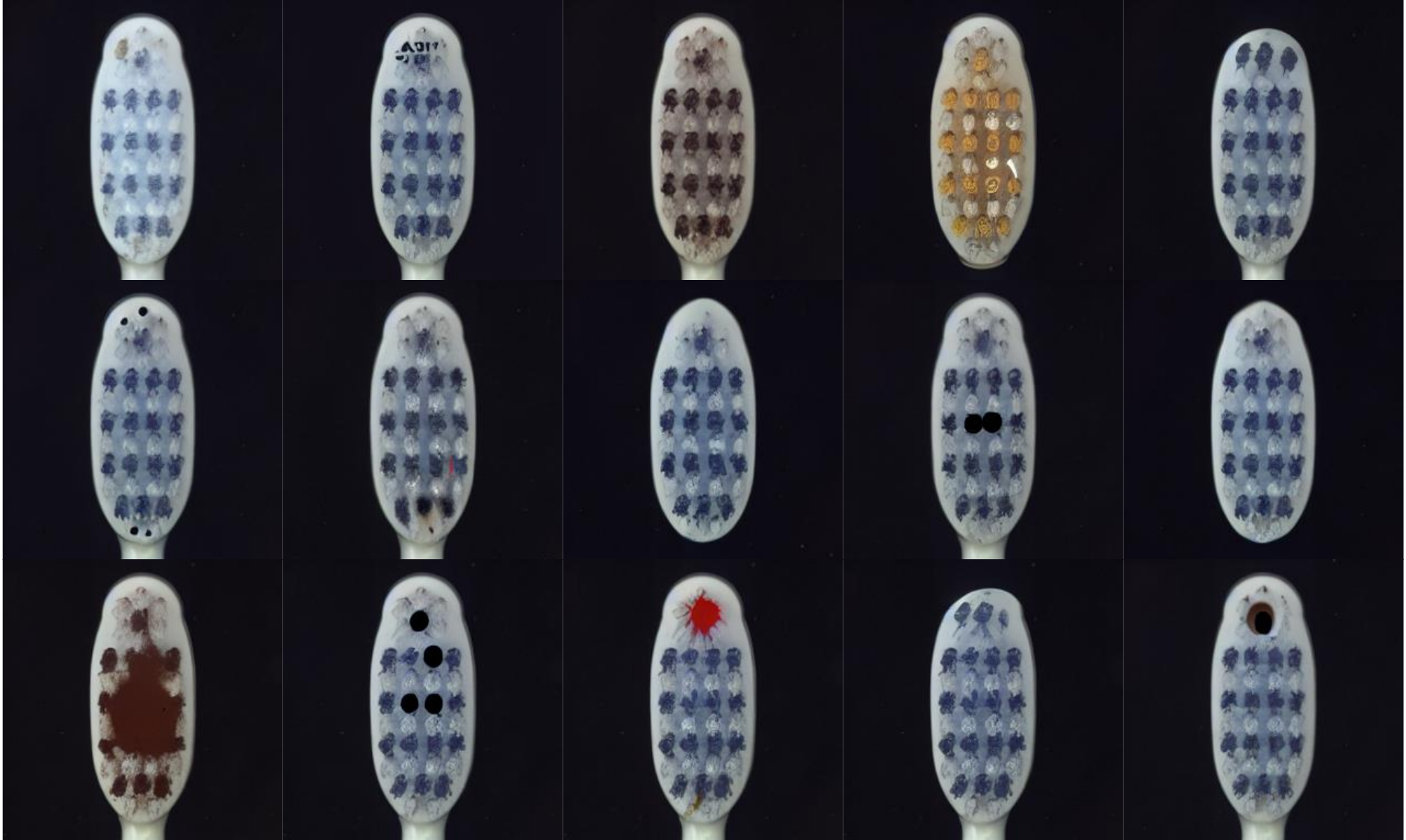} 
  \caption{Additional qualitative results: Synthetic toothbrush anomaly}
\end{figure*}

\vspace{5cm}

\begin{figure*}[b]
    \includegraphics[width=\textwidth, page=9 ,trim={1cm 0cm 1cm 0cm},clip]{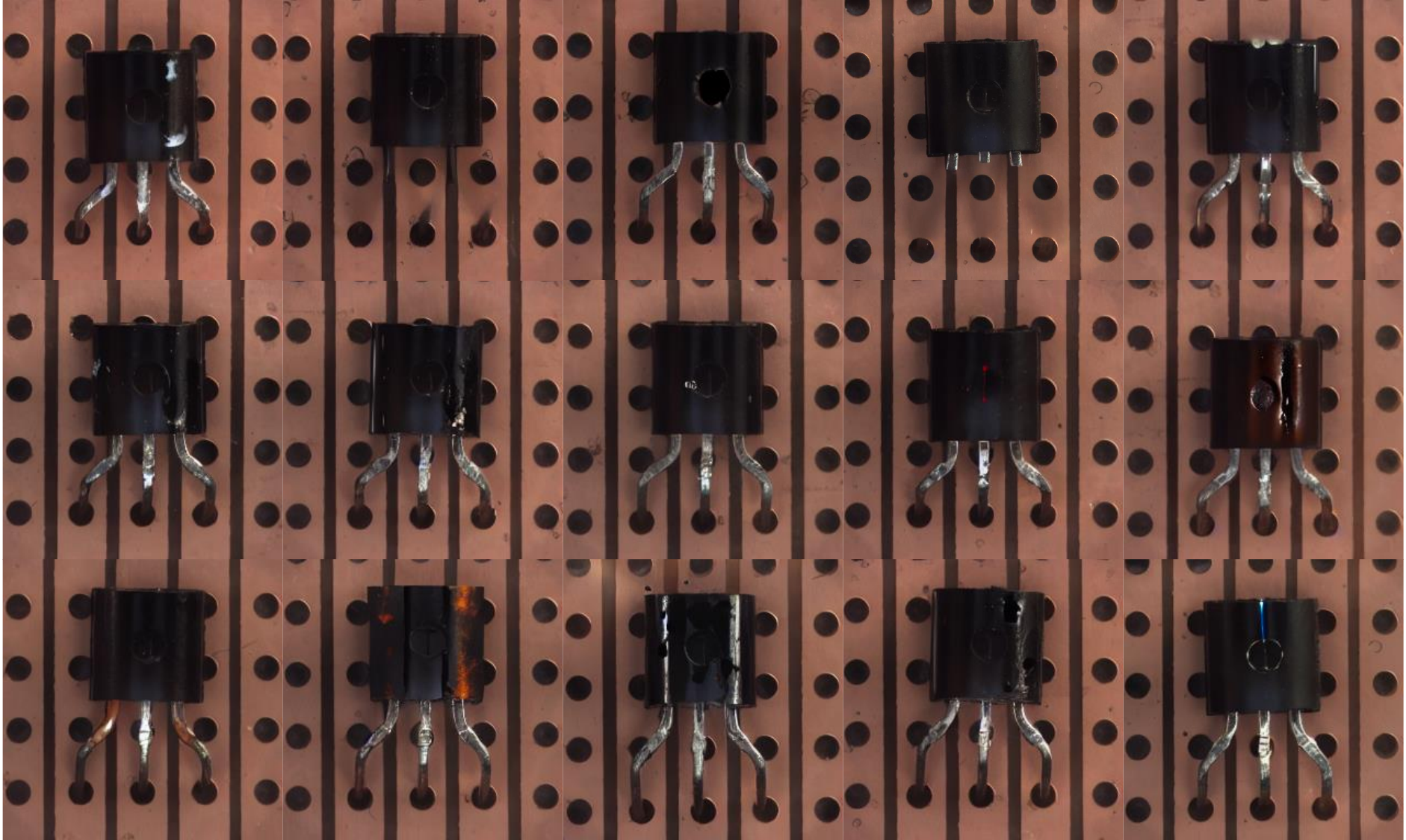} 
  \caption{Additional qualitative results: Synthetic transistor anomaly}
\end{figure*}

\begin{figure*}[b!]
    \includegraphics[width=\textwidth, page=10 ,trim={0cm 0cm 0cm 0cm},clip]{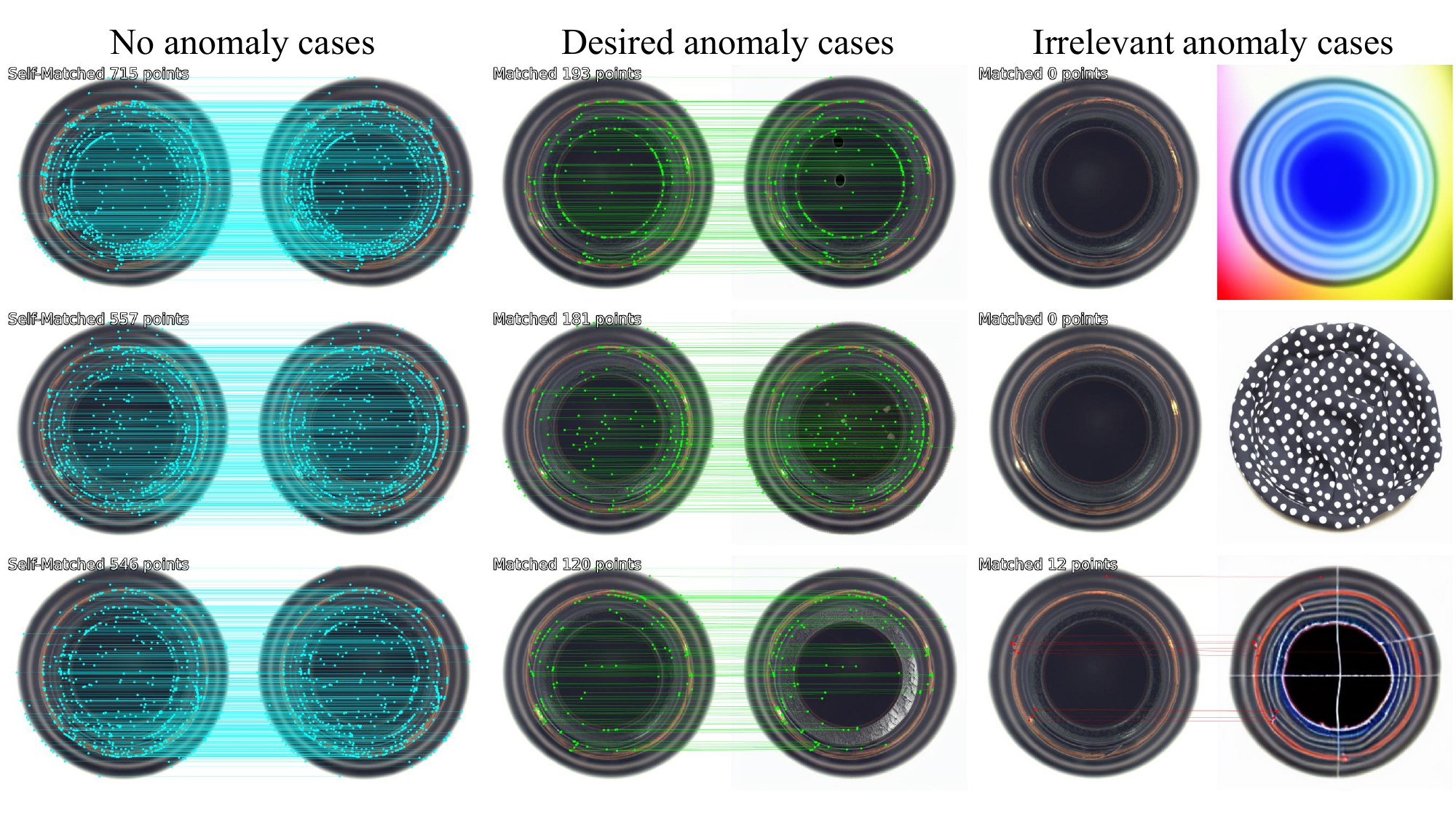} 
  \caption{Additional matching point-based filtering result: Bottle}
\end{figure*}

\vspace{5cm}

\begin{figure*}[t!]
    \includegraphics[width=\textwidth, page=11 ,trim={0cm 0cm 0cm 0cm},clip]{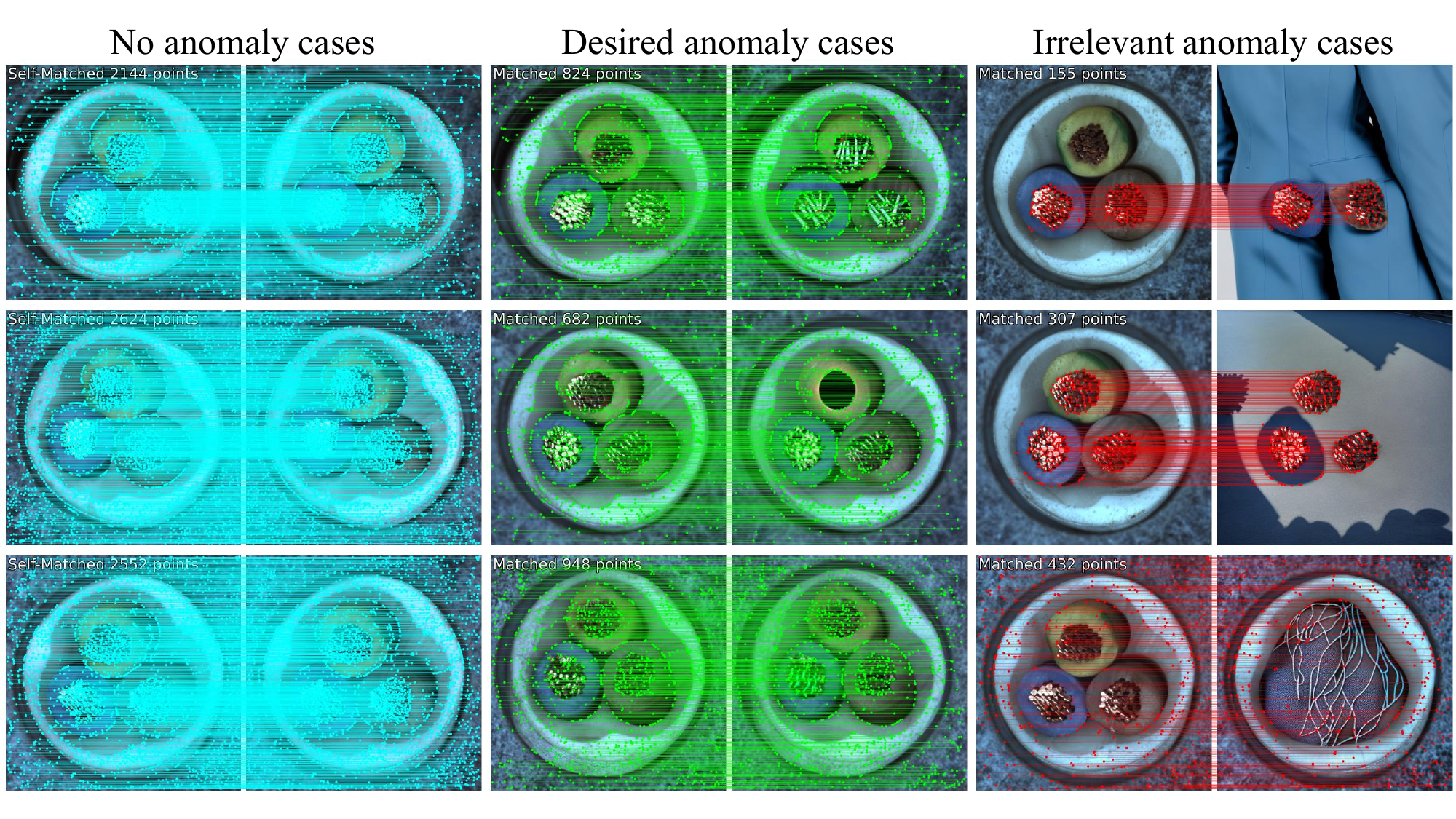} 
  \caption{Additional matching point-based filtering result: Cable}
\end{figure*}

\vspace{5cm}

\begin{figure*}[b!]
    \includegraphics[width=\textwidth, page=12 ,trim={0cm 0cm 0cm 0cm},clip]{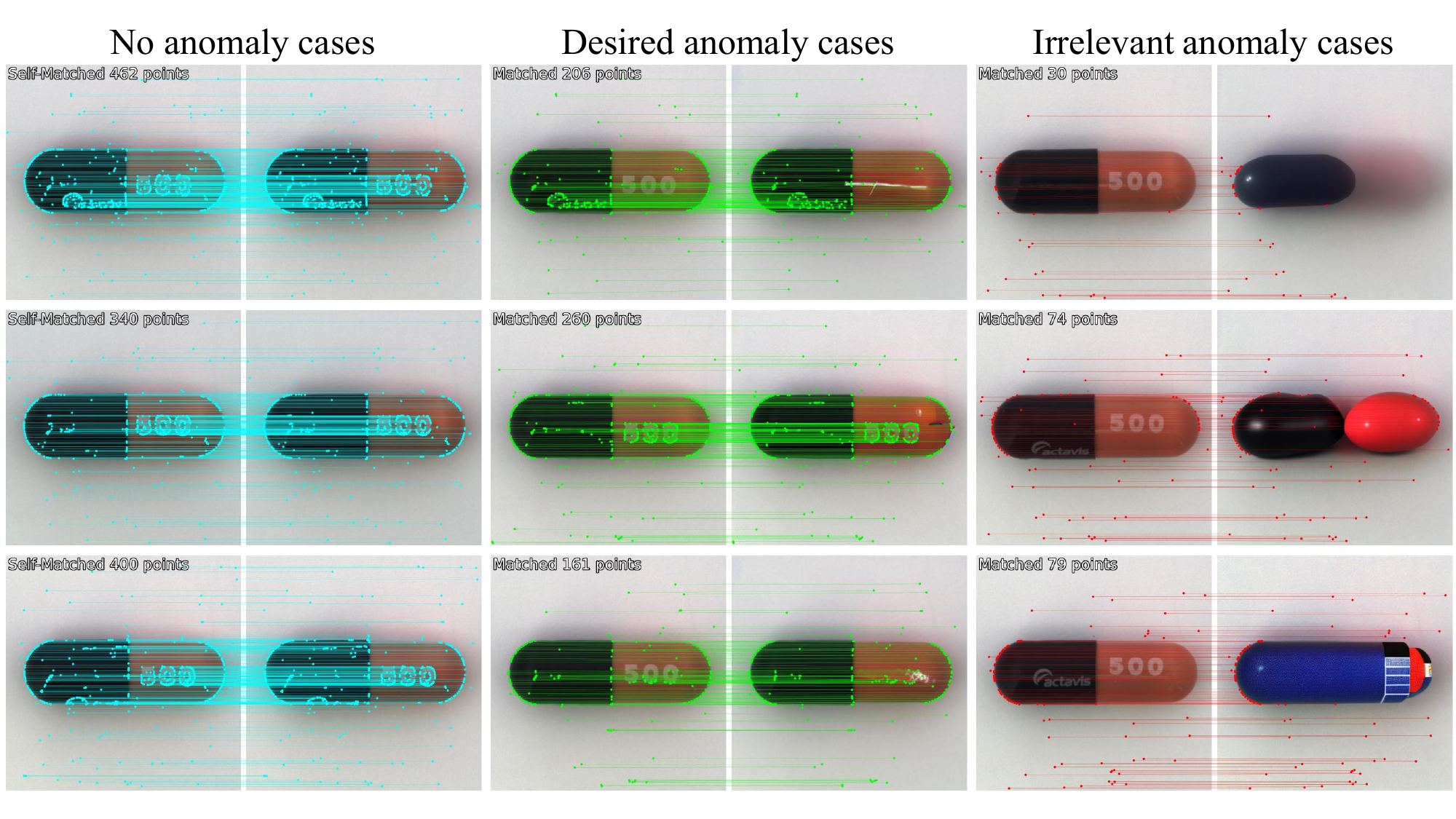} 
  \caption{Additional matching point-based filtering result: Capsule}
\end{figure*}

\begin{figure*}[t!]
    \includegraphics[width=\textwidth, page=13 ,trim={0cm 0cm 0cm 0cm},clip]{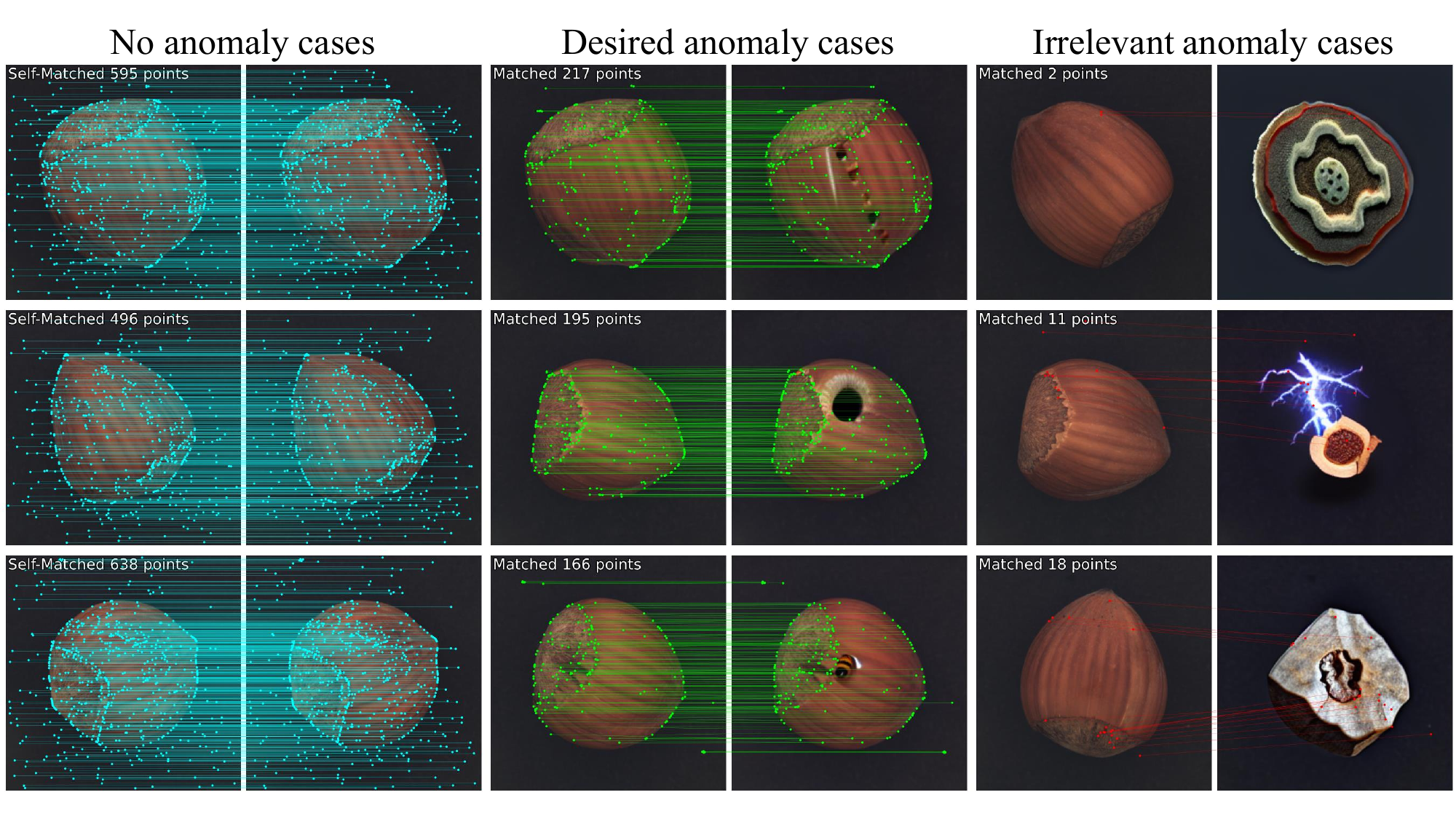} 
  \caption{Additional matching point-based filtering result: Hazelnut}
\end{figure*}

\vspace{5cm}

\begin{figure*}[b!]
    \includegraphics[width=\textwidth, page=14 ,trim={0cm 0cm 0cm 0cm},clip]{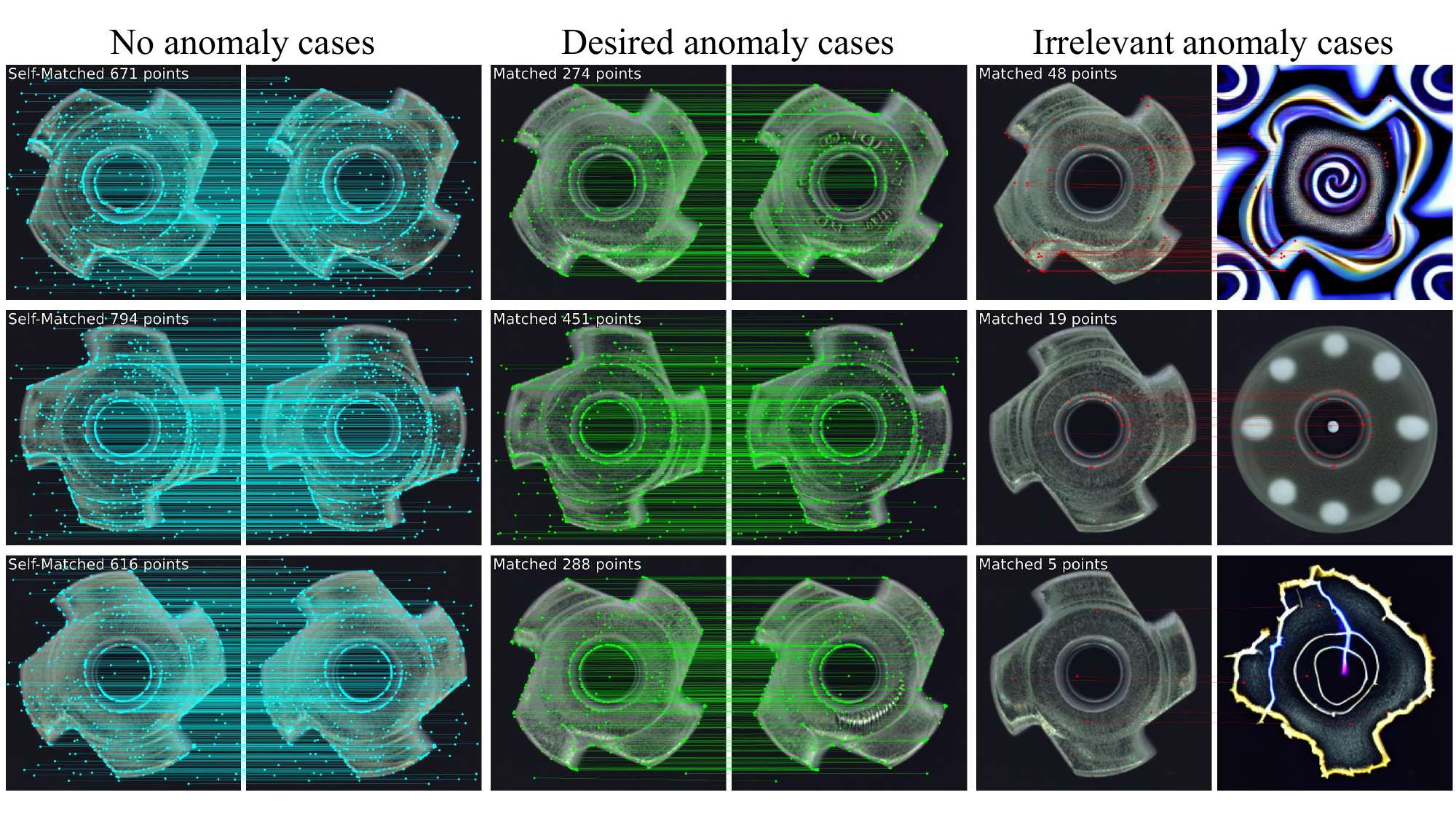} 
  \caption{Additional matching point-based filtering result: Metal nut}
\end{figure*}

\begin{figure*}[t!]
    \includegraphics[width=\textwidth, page=15 ,trim={0cm 0cm 0cm 0cm},clip]{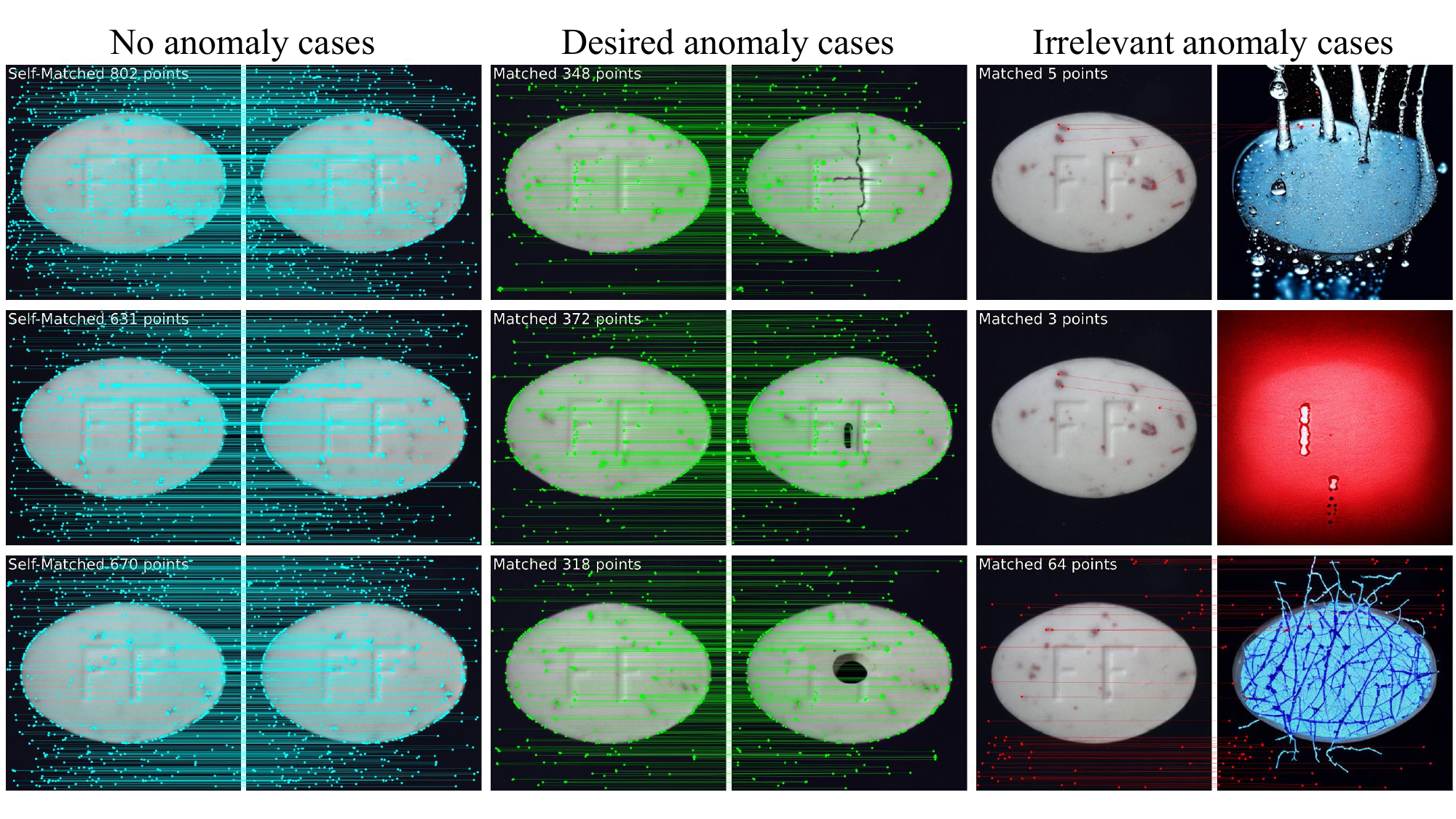} 
  \caption{Additional matching point-based filtering result: Pill}
\end{figure*}

\vspace{5cm}

\begin{figure*}[b!]
    \includegraphics[width=\textwidth, page=16 ,trim={0cm 0cm 0cm 0cm},clip]{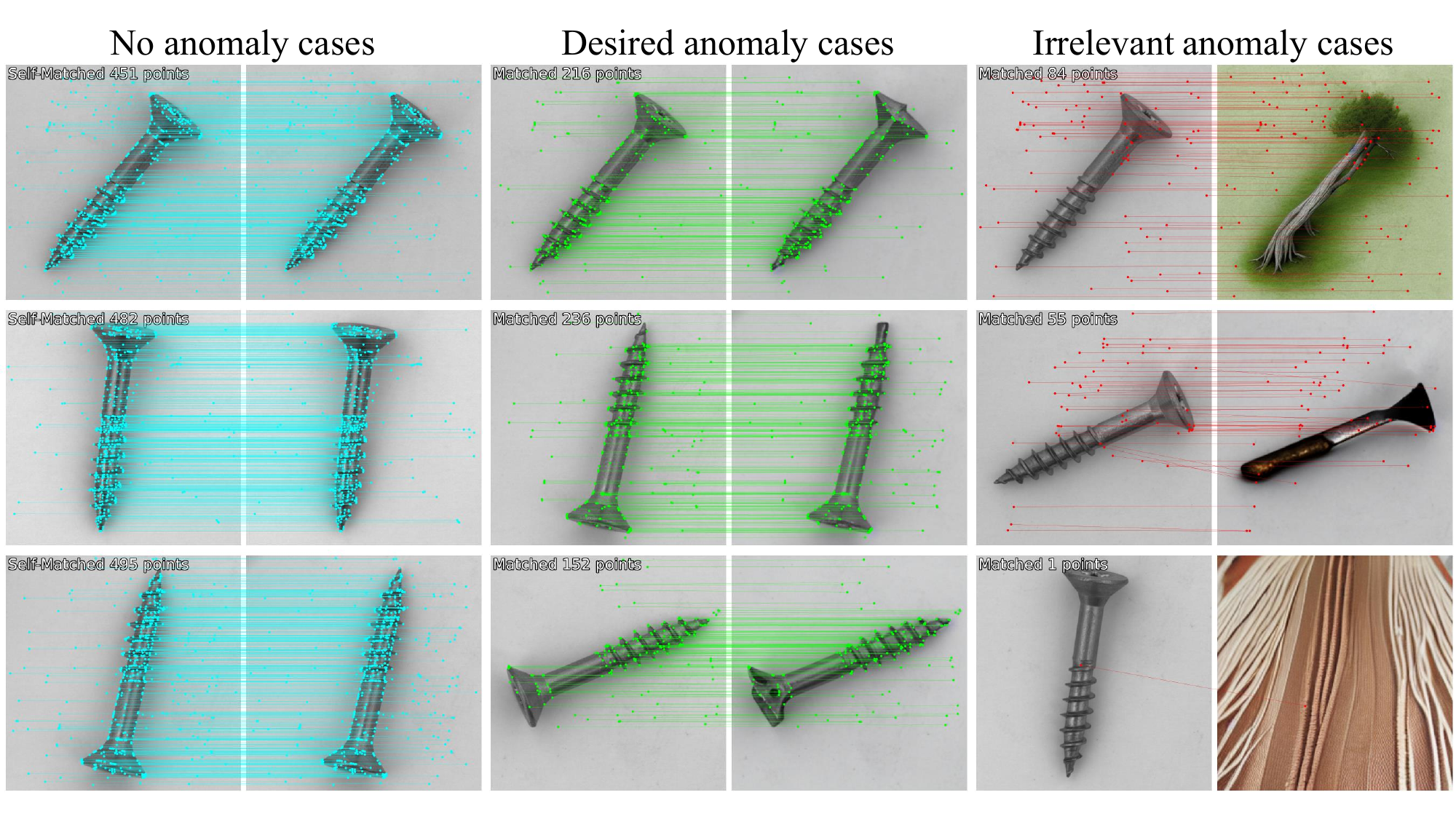} 
  \caption{Additional matching point-based filtering result: Screw}
\end{figure*}

\begin{figure*}[t!]
    \includegraphics[width=\textwidth, page=17 ,trim={0cm 0cm 0cm 0cm},clip]{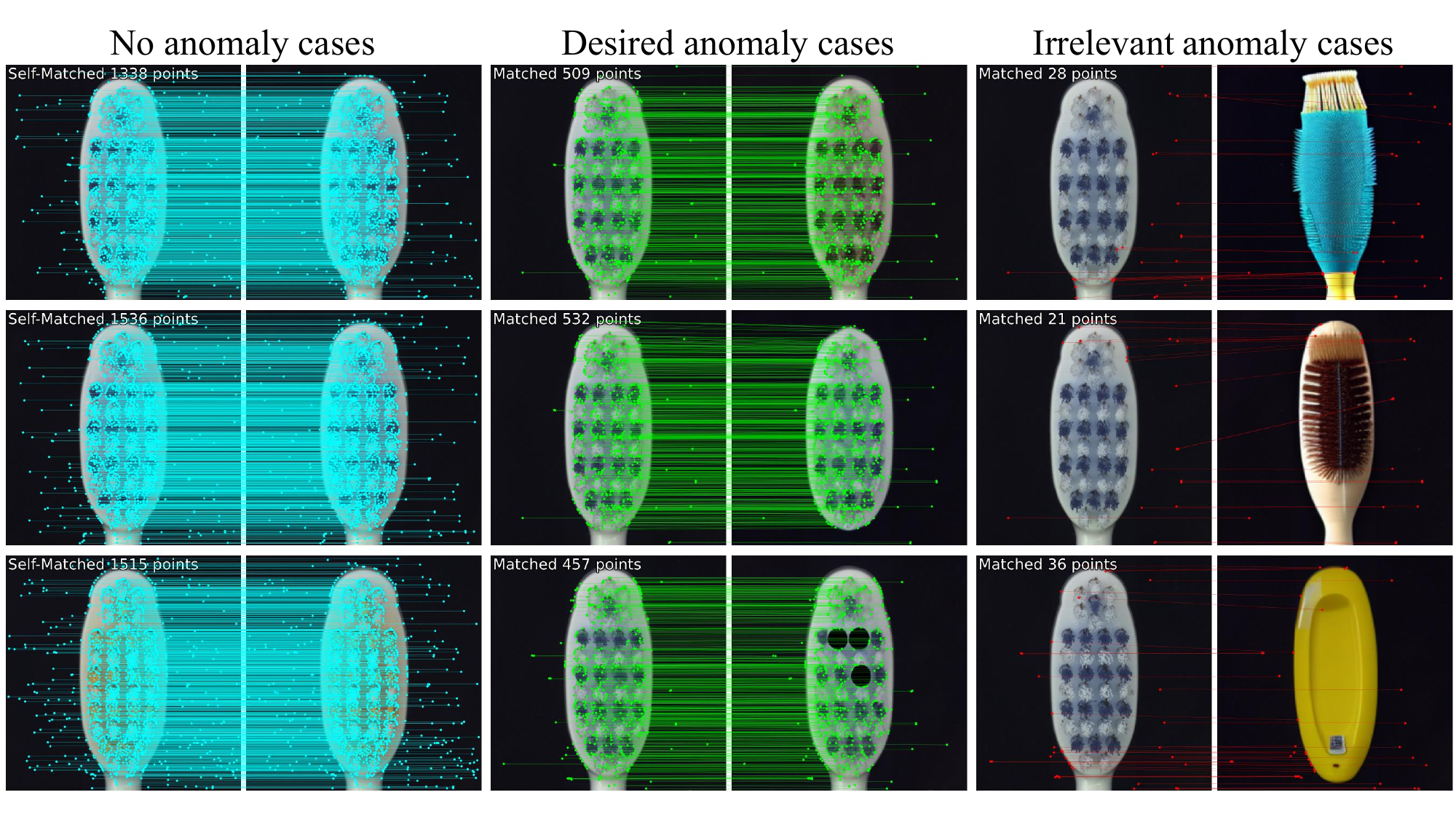} 
  \caption{Additional matching point-based filtering result: Toothbrush}
\end{figure*}

\vspace{5cm}

\begin{figure*}[b!]
    \includegraphics[width=\textwidth, page=18 ,trim={0cm 0cm 0cm 0cm},clip]{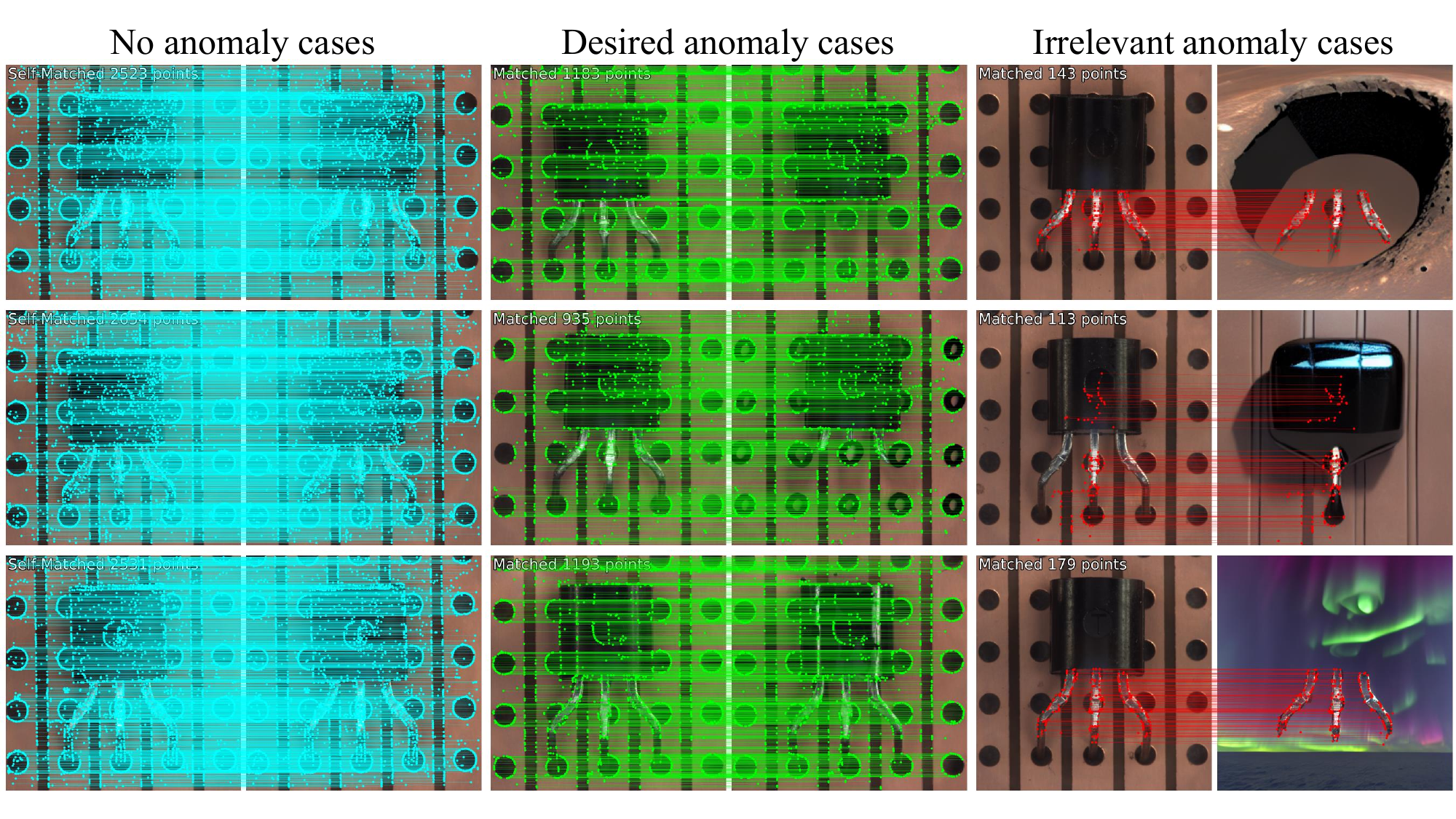} 
  \caption{Additional matching point-based filtering result: Transistor}
\end{figure*}

\end{document}